\documentclass[10pt,journal,compsoc]{IEEEtran}

\ifCLASSOPTIONcompsoc
  \usepackage[nocompress]{cite}
\else
  \usepackage{cite}
\fi

\ifCLASSINFOpdf
\usepackage[pdftex]{graphicx}
\DeclareGraphicsExtensions{.pdf,.jpeg,.png}
\DeclareGraphicsExtensions{.eps}
\usepackage{epstopdf}
\else
\fi

\ifCLASSOPTIONcompsoc
\usepackage[caption=false,font=footnotesize,labelfont=sf,textfont=sf]{subfig}
\else
\usepackage[caption=false,font=footnotesize]{subfig}
\fi
\usepackage{booktabs}
\usepackage{amssymb}

\usepackage{mathrsfs}
\usepackage{times}
\usepackage{epsfig}
\usepackage{graphicx}
\usepackage{amsmath}
\usepackage{cite}
\usepackage{enumerate}
\usepackage{cases}
\usepackage{multirow}
\usepackage{verbatim}
\usepackage{amssymb}
\usepackage{CJK}
\usepackage{algorithm}
\usepackage{algorithmicx}
\usepackage{algpseudocode}

\usepackage{color}
\usepackage{bm}
\usepackage{booktabs}
\usepackage{colortbl}
\usepackage{float}
\usepackage{ragged2e}  
\usepackage{caption}
\usepackage{csquotes}
\usepackage[breaklinks=true,bookmarks=false]{hyperref}
\usepackage{breakurl}

\newenvironment{packed_itemize}{
	\vspace{-0.15cm}\begin{itemize}
		\setlength{\itemsep}{1pt}
		\setlength{\parskip}{0pt}
		\setlength{\parsep}{0pt}
	}{\end{itemize}}

\begin{document}

\title{Flexible Piecewise Curves Estimation \\ for Photo Enhancement}

\author{Chongyi Li, Chunle Guo, Qiming Ai, Shangchen Zhou, and Chen Change Loy,~\IEEEmembership{Senior Member,~IEEE}

\thanks{C. Li, Q. Ai, S. Zhou, and C. C. Loy are the School of Computer Science and Engineering, Nanyang Technological University (NTU), Singapore  (e-mail: chongyi.li@ntu.edu.sg, qai001@e.ntu.edu.sg, sczhou@ntu.edu.sg, and ccloy@ntu.edu.sg).}
\thanks{C. Guo is with the College of Computer Science, Nankai University, Tianjin, China  (e-mail: guochunle@nankai.edu.cn).}
\thanks{C. C. Loy is the corresponding author.}
}


\IEEEtitleabstractindextext{%
\justify  

\begin{abstract}
	\label{sec:Abstrat}
	%
	This paper presents a new method, called FlexiCurve, for photo enhancement.
	Unlike most existing methods that perform image-to-image mapping, which requires expensive pixel-wise reconstruction, FlexiCurve takes an input image and estimates global curves to adjust the image. The adjustment curves are specially designed for performing piecewise mapping, taking nonlinear adjustment and differentiability into account.
	To cope with challenging and diverse illumination properties in real-world images, FlexiCurve is formulated as a multi-task framework to produce diverse estimations and the associated confidence maps.  These estimations are adaptively fused to improve local enhancements of different regions.
	Thanks to the image-to-curve formulation, for an image with a size of 512$\times$512$\times$3, FlexiCurve only needs a lightweight network (150K trainable parameters) and it has a  fast inference speed (83FPS on a single NVIDIA 2080Ti GPU). 
	The proposed method improves efficiency without compromising the enhancement quality and losing details in the original image.
	The method is also appealing as it is not limited to paired training data, thus it can flexibly learn rich enhancement styles from unpaired data.  
	Extensive experiments demonstrate that our method achieves state-of-the-art performance on photo enhancement quantitively and qualitatively. 
\end{abstract}

}

\maketitle

\IEEEdisplaynontitleabstractindextext

\IEEEpeerreviewmaketitle


\section{Introduction}
\label{sec:Introduction}

Image enhancement enjoys remarkable progress thanks to deep learning. Despite the impressive performance of current methods, they suffer from some limitations.
In general, large networks are required to cope with pixel-wise reconstruction,  which inevitably leads to high memory footprint and long inference time due to massive parameter space.
Improving the efficiency using shallow networks will compromise the ability of preserving or enhancing local details. Shallow networks may even generate artifacts and artificial colors due to their limited capacity.
In addition, many existing methods \cite{DSLR,DBLNet} require paired training data and thus they can only produce a fixed adjustment style.  
Relying on paired training data increases the risk of overfitting on specific data and limits the flexibility of these models to meet diverse user requirements.
Clearly, trading-off efficiency, enhancement performance, and flexibility is still an open research problem in photo enhancement.

To address the aforementioned problems, we draw inspirations from existing photo editing tools such as Adobe Photoshop.
In these tools, users usually remap the tonality of an image by adjusting a curve that specifies a function from input level to output level of a specific color channel. 
For instance, one can adjust the highlight of the image by moving a point in the top portion of the curve; or adjust the shadows by changing a point in the bottom section of the curve. Moving a point up and down lightens or darkens the tonal area while dragging a point left to right changes the contrast.
Curve adjustment can be made more flexible by adding more control points (or knot points) to adjust different tonal areas.

\begin{figure*}[t]
	\centering
	\includegraphics[width=1\linewidth]{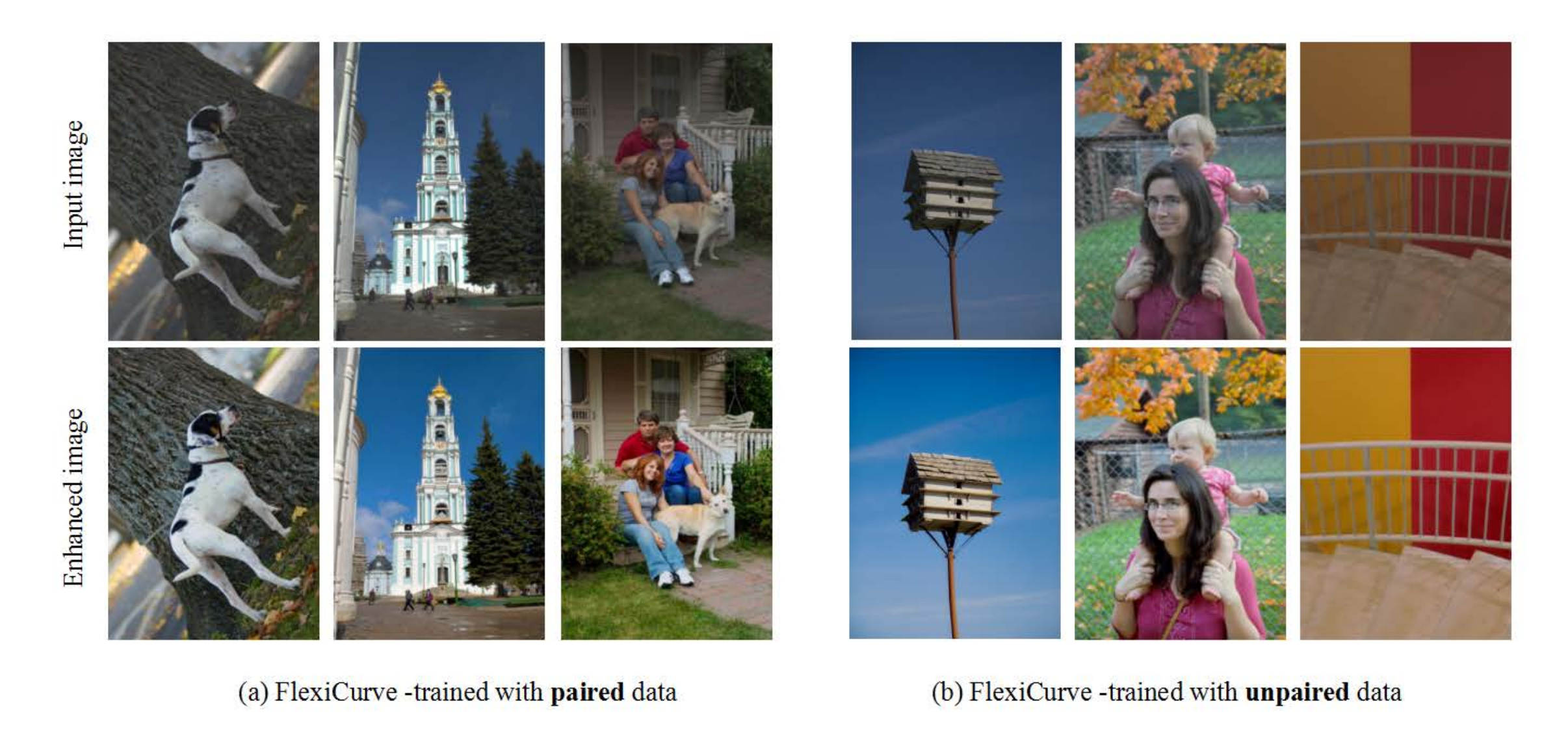}
	\caption{Visual examples by FlexiCurve. The first row represents the low-quality images sampled from Adobe 5K dataset \cite{Adobe5K}. The second row represents the corresponding results by the proposed FlexiCurve, where (a) represents the results enhanced by FlexiCurve trained with paired data and (b) represents the results enhanced by FlexiCurve trained with unpaired data. The FlexiCurve can deal with global tone and local contrast well and does not introduce over-/under-enhancement regions regardless of paired or unpaired training data.}
	\label{fig:teaserfigure}
\end{figure*}

We found such curve adjustment properties appealing for deep learning-based photo enhancement. Instead of generating an entire image as an output, a curve estimation network only needs to estimate a handful of curve parameters. Enhancing an image can be achieved by just curve mapping, which significantly simplifies the enhancement network and makes the whole process more efficient than conventional pixel-wise reconstruction. Besides, curve adjustment can preserve details and inherent relations of neighbouring pixels well through the direct mapping of input level to the output level.

Despite the attractive properties of curve adjustment, devising a deep curve estimation network for photo enhancement is non-trivial.
A na\"{i}ve global curve treats the whole image homogeneously while estimating a transformative coefficient for every single pixel is too expensive. The challenge is to devise a network and the corresponding output space to cater for piecewise curves, so as to balance the flexibility of curves and efficiency. The process is akin to placing control points on the right position on a curve and moving them in a correct direction and level. 
Another challenge is to cope with diverse properties and illuminations in different regions of an image. Piecewise curve adjustment is still global, and thus the results may experience over enhancement in the bright regions, under enhancement in the dark regions, or producing unnatural artifacts. 

In this work, we formulate a new method called FlexiCurve, which enjoys the benefits of curve adjustment yet capable of overcoming the aforementioned challenges.
FlexiCurve takes an RGB image as input to estimate the corresponding adjustment curves. To allow flexibility in curve estimation, we adopt a multi-task network and design the output space to encompass both the knot points to constrain the adjustment range of each sub-function of a piecewise curve and the associated curve parameters to control the curvature. 

To cope with diverse and heterogeneous regions in images, we further allow the network to produce multiple adjustment curves for curve mapping. This gives us multiple solutions that are complementary to each other - one of the solutions may perform very well on bright region, while another solution may emphasize more on the shadow region. Fusing these solutions allows us to handle both global and local properties such as color, textures, and contrast well.
The fusion is made possible by using spatial-adaptive confidence maps, which are also the outputs of the FlexiCurve.  
It is noteworthy that further constraints can be imposed on the multiple solutions so that each of them can serve as a standalone and sensible output but with different styles. We also show an in-depth theoretical analysis on how to ensure the solutions to lie on a manifold that fulfils solution space constraints.
Such mechanism allows our method to offer users with additional flexibility to explore different enhancement options on different spatial regions, which is impossible with existing methods that can only provide a single output. 

Thanks to the succinct design, the proposed FlexiCurve only consists of 153,488 trainable parameters, which is about 191 times fewer than popular network structures such as U-Net \cite{UNet} (31,042,434 trainable parameters), which is used in~\cite{Chen18,Enhancer} for image enhancement. The inference speed of FlexiCurve is about 83/1FPS on an
NVIDIA 2080Ti GPU or an Intel Core i9-10920X
CPU@3.5GHz CPU for an image with a size of 512$\times$512$\times$3, repectively.
The training of the proposed method requires no paired data, unlike existing approaches~\cite{DSLR,DBLNet}. This is made possible by introducing specially designed content loss and adversarial loss. Our method can learn from paired data if available. 
Some representative results of our method are shown in Figure \ref{fig:teaserfigure}, suggesting the capability of FlexiCurve in returning visually pleasing results with and without using paired data for training.

We summarize the contributions of this work as follows:
\begin{packed_itemize}
	\item We present a novel approach that performs photo enhancement by piecewise curves estimation through a deep network. Unlike existing methods that spend expensive computation for pixel-wise reconstruction, the new notion of curve adjustment enables parameter-efficient network structure, real-time inference, and good details preservation and enhancement.
	\item We explain how piecewise adjustment curves can be designed as the network's output. We show that a flexible curve can be realized through knot points and nonlinear curve parameters, which collectively perform piecewise and nonlinear adjustment while being differentiable in the process of gradient back-propagation.
	\item We devise a theoretically sound method to cope with diverse and heterogeneous regions in images. The proposed network can generate multiple plausible results that can be fused with spatial-adaptive confidence maps, producing visually pleasing results both globally and locally. 
\end{packed_itemize}

\section{Related Work}
\label{sec:Related_Work}

\noindent
\textbf{Photo Enhancement.}
Early works on photo enhancement mainly perform color adjustment, ranging from example-based methods \cite{Erilk,Hwang14} to retrieval-based methods \cite{Lee16}. 
For example, Reinhard et al. \cite{Erilk} proposed to impose one image's color characteristics  on another based on statistical analysis. Lee et al. \cite{Lee16} learned a content-specific style ranking using semantic and style similarity and transferred  the global color and tone of the chosen example images to the input image. 
These methods may suffer from potential visual artifacts due to improper exemplar.  
Unlike these methods that transform an input image towards exemplar images, our method can enhance a given photo towards a similar distribution with a reference image collection.

Learning-based methods are popular for photo enhancement. Earlier work such as Yan et al. \cite{Yan16} learn specific enhancement styles given the features of color and semantic context. Learning-based photo enhancement solutions spread from  Convolutional Neural Network (CNN)-based, Generative Adversarial Network (GAN)-based, to Reinforcement Learning (RL)-based methods. Specifically,
CNNs were employed to approximate image processing operators \cite{Chen17}, reconstruct latent images \cite{DSLR}, estimate affine transforms \cite{DBLNet}, or learn parametric filters \cite{Moran20} by supervised learning on paired training data. 
For instance, Gharbi et al. \cite{DBLNet} trained CNNs to predict the coefficients of a locally-affine model in bilateral space, then apply these coefficients to photo enhancement. 
Unpaired adversarial learning based on the GAN framework has been used for photo enhancement. For example, a two-way GAN was proposed to enhance images by pixel-wise reconstruction \cite{Enhancer}. 
Deng et al. \cite{Deng18} proposed an EnhanceGAN for automatic image enhancement, which requires binary labels on image aesthetic quality. 
Ni et al. \cite{Ni2020TIP} proposed to learn image-to-image mapping from a set of images with desired characteristics in an unsupervised manner for photo enhancement.
More recently, Park et al. \cite{Park18} proposed an RL-based color enhancement method to learn the optimal global enhancement sequence of retouching actions. 
Similarly, a white-box photo post-processing framework was proposed \cite{WhiteBox}, which applies different retouching operations on an input image based on RL decision. 
Besides, there are several specific research areas for image enhancement such as image super-resolution \cite{SRCNN,Zhang20}, image denoising \cite{Liu14,Guo19}, image deblurring \cite{Pan18,Pan20}, low-light/underexposed image enhancement \cite{Chen18,Jiang2019,Wang2019,Yang20,Zero-DCE,Xu2020CVPR}, high dynamic range (HDR) image reconstruction \cite{Lee18,Liu20}, etc. 

In comparison to existing deep learning-based photo enhancement models, we deal with the problem from a different view. 
Instead of performing image-to-image mapping, we perform image-to-curves mapping then use the piecewise curves for adjustment. The whole process can be trained end-to-end as we design the curves to be differentiable in the process of gradient back-propagation. The formulation allows FlexiCurve to enjoy good enhancement performance, lightweight network structure, and fast inference speed. We further propose the one-to-many framework to produce multiple plausible solutions. This mechanism is theoretically sound. It not only improves the robustness in coping with diverse illumination in local regions of images, but also offers the flexibility for user to explore different enhancement options. 

\noindent
\textbf{Image-to-Image Translation.}
Image-to-image translation aims at mapping images in a given class to an analogous image of a different class. Popular methods include Pix2Pix \cite{CGANs}, Cycle-GAN \cite{CycleGANs}, few-shot unsupervised image-to-image translation \cite{unsupervisedGANs},  semantic photo manipulation \cite{Bau19}, etc. 
%
There are two important differences between image-to-image translation and our work: 1) instead of generating an image directly, our method first predicts piecewise curve parameters and then applies them to a given image by curve mapping, which effectively preserves the relations of neighboring pixels, and 2) unlike traditional image-to-image translation that only produces a single translation for each input, our method can produce diverse enhancement results.

Our method is also related to recent curve parameter estimation methods DeepLPF \cite{Moran20}, Chai et al. \cite{Chai2020}, CURL \cite{CURL}, and Zero-DCE \cite{Zero-DCE}. In comparison to DeepLPF \cite{Moran20}, there are three important differences: 1)  unlike DeepLPF that estimates the parameters of three specific filters (i.e., elliptical filter, graduated filter, and polynomial filter) for enhancing images, our method automatically learns optimal parameters of specially designed piecewise curves, which does not limit in the fixed filters that cannot cover all cases, 2) different from DeepLPF that requires paired training data, our method allows both paired and unpaired training, thus it can flexibly learn different enhancement styles, and 3) our method allows  user to further adjust the automatically produced results according to their preference. 

Chai et al. \cite{Chai2020} treated the problem of color enhancement as an image translation task. The translation is performed by using a global parameterized quadratic transformer that includes the fixed 1$\times$10 quadratic color basis vector. Unlike \cite{Chai2020} that designs a per-pixel color transformation, we propose the specially designed piecewise curves to balance the flexibility of curves and efficiency. Moreover, our method not only handles the color enhancement well, but also effectively improves the contrast and saturation.

CURL \cite{CURL} is proposed for global image enhancement while our method can perform both global and local enhancement by spatial-adaptive fusion and is not limited in paired training data, which flexibly deal with challenging cases. 
CURL adopts piecewise linear scaling for pixel values while we nonlinearly adjust the curvature of each sub-function of a curve, thus improving the flexibility of curve mapping. Morveover, we  adaptively integrate multiple curves estimation by confidence maps. Besides, our multi-task parameter estimation network does not require multi-color space transformation, fully connected layers, and complex loss functions thanks to the effective designs of piecewise curves estimation and spatial-adaptive fusion. 

Compared with the recent Zero-DCE \cite{Zero-DCE}, which is designed for low-light image enhancement based on pixel-wise curve parameters estimation, our method estimates piece-wise curve parameters and enjoys the good flexibility in adjusting the curvature in each piece. Moreover, our method does not require complex constraints in the training phase and can learn diverse enhancement styles.

\begin{figure*}[t]
	\begin{center}
		\begin{tabular}{c@{ }}
			\includegraphics[width=1\textwidth]{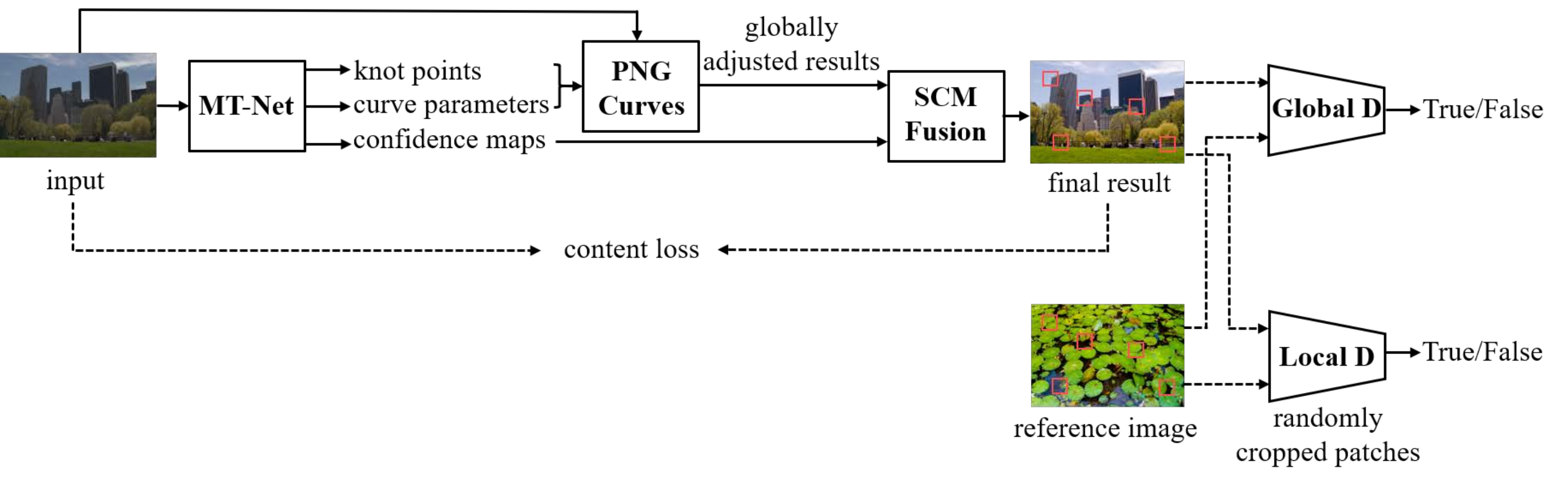}\\
			(a) FlexiCurve trained with unpaired data \\
			\\
			\includegraphics[width=1\textwidth]{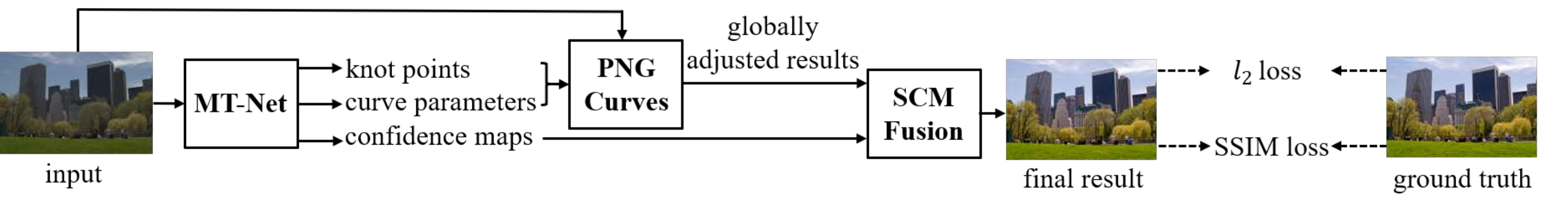}\\
			(b) FlexiCurve trained with paired data  \\
		\end{tabular}
	\end{center}
	\caption{An overview of the FlexiCurve framework. The input image is first fed to the multi-task network (MT-Net) for estimating a set of knot points, curve parameters, and confidence maps. The knot points and curve parameters define the piecewise nonlinear global curves (PNG Curves), which are used to adjust the level of the RGB channels of an input image. Multiple curves are generated simultaneously to provide us with multiple globally adjusted results. The final result is achieved by blending such intermediate results via spatial-adaptive confidence map fusion (SCM Fusion). (a) FlexiCurve trained with unpaired data, where global and local discriminators are employed to distinguish whether the final image or randomly cropped patches are `real' or `fake', respectively. To preserve the original image content, a content loss is employed in the training phase. (b) FlexiCurve trained with paired data, where the $\ell_{2}$ and SSIM losses are employed for supervised training.}
	\label{fig:pipeline}
\end{figure*}


\section{Methodology}
\label{sec:Method}

We present the overview of FlexiCurve framework with unpaired training data in Figure \ref{fig:pipeline}(a). The framework can be easily extended to supervised learning if paired data are available. The  FlexiCurve framework with paired training data is presented in Figure \ref{fig:pipeline}(b)
As illustrated, the proposed framework is specially designed to allow flexibility in curve estimation.
This is achieved through a multi-task network (MT-Net), which estimates a set of optimal knot points, curve parameters, and the associated confidence map for a curve. 
The framework then separately maps the pixels in the RGB channels of a given image by piecewise global adjustment curves defined by knot points and curve parameters. 

To cope with diverse properties and illuminations of different regions in the input image, the framework produces a set of piecewise nonlinear global curves (PNG Curves), each of which enhances regions in the image differently.
Given multiple sets of global adjustment curves, our approach generates multiple globally adjusted results (each result consists of three adjusted RGB channels) and the corresponding confidence maps. The final result can be achieved by fusing these globally adjusted results spatially using the confidence maps. We call this process as spatial-adaptive confidence map fusion (SCM Fusion).

During the unpaired training phase, we adopt a content loss to preserve the semantic content of the input image and introduce the global and local discriminators to capture the desired manifold of the reference image collection.  For paired training, we remove the content loss and adversarial losses  and only use the $\ell_{2}$ and SSIM  \cite{Zhao2017} losses to train this framework. 
We next detail the key components in our framework, namely PNG Curves, MT-Net, SCM Fusion, and loss functions.

\subsection{Piecewise Nonlinear Global Curve}
\label{sec:PNG-Curve}
Drawing inspiration from previous works \cite{Zero-DCE,CURL}, we design PNG Curve that allows 1) adjustment of input pixels in a piecewise manner, which constrains the adjustment range of each sub-function of a piecewise curve by some knot points, 2) nonlinear mapping in each piece, which controls the curvature by estimating the associated curve parameters  and 3)  end-to-end training as the curve is differentiable. The formulation of PNG Curve can be expressed as:
\begin{equation}
\label{equ_1}
V=k_{0}+\sum^{M-1}_{m=0}(k_{m+1}-k_{m})\mathbf{G}(I;m,\alpha_{m}),
\end{equation}
\begin{equation}
\label{equ_1_1}
\mathbf{G}(I;m,\alpha_{m})=\mathbf{F}_{c}^{n}(\mathbf{F_\delta}(M-m);\alpha_{m}),
\end{equation}
where  $I\in$[0,1] and $V$ are the input channel and globally adjusted channel, respectively. The variable $k_m$  denotes the value of $m$th knot point and $k_{m+1}-k_{m}$ constrains the adjustment ranges in the corresponding piece. The total number of pieces is represented by $M$, thus leading to $M+1$  knot points. The curvature of nonlinear curve in each piece is controlled by $\alpha$ $\in$[-1,1] and $\alpha_{m}$ is the nonlinear curve parameter of the $m$th piece. Both $k$ and $\alpha$ are learned by the MT-Net. The $m$th piecewise curve  is represented by $(k_{m+1}-k_{m})\mathbf{G}(I;m,\alpha_{m})$, which can be further derived as in Eq.~\eqref{equ_1_1}, where $\mathbf{F}_{c}^{n}$ controls the nonlinearity in each piece, which can obtain a larger curvature by iteratively applying the basic curve $\mathbf{F_{c}}$ for $n$ times.  
The basic curve $\mathbf{F}_{c}$ can be expressed as:
\begin{equation}
\label{equ_2}
\mathbf{F}_{c}(x;\alpha)=x+\alpha x(1-x),
\end{equation}
where $x$ is the input of $\mathbf{F}_{c}$. Thus, $\mathbf{F}_{c}^{n}$ can be expressed as:
\begin{equation}
\label{equ_3}
\mathbf{F}_{c}^{n}=\mathbf{F}_{c}^{n-1}+\alpha\mathbf{F}_{c}^{n-1}(1-\mathbf{F}_{c}^{n-1}),
\end{equation}
where $n$ is the number of iteration, which can further increase or decrease the curvature. More flexible curvature is important for challenging cases, such as extremely dark or over-exposed regions. We set the number of iteration $n$ to 4 for the trade-off between performance and efficiency. We will investigate the effect of iteration number in the ablation study. The function $\mathbf{F_\delta}$ in Eq. \eqref{equ_1_1} makes the PNG Curve successive and differentiable. It can be expressed as:
\begin{equation}
\label{equ_4}
\mathbf{F_\delta}(y)=\left\{
\begin{aligned}
0 &  & y<0 \\
y &  & 0\leq y \leq 1 \\
1 &  &  y>1
\end{aligned},
\right.
\end{equation}
where  $y$ denotes the input of the function $\mathbf{F_\delta}$. 

\begin{figure*}[t]
	\centering
	\centerline{\includegraphics[width=0.8\linewidth]{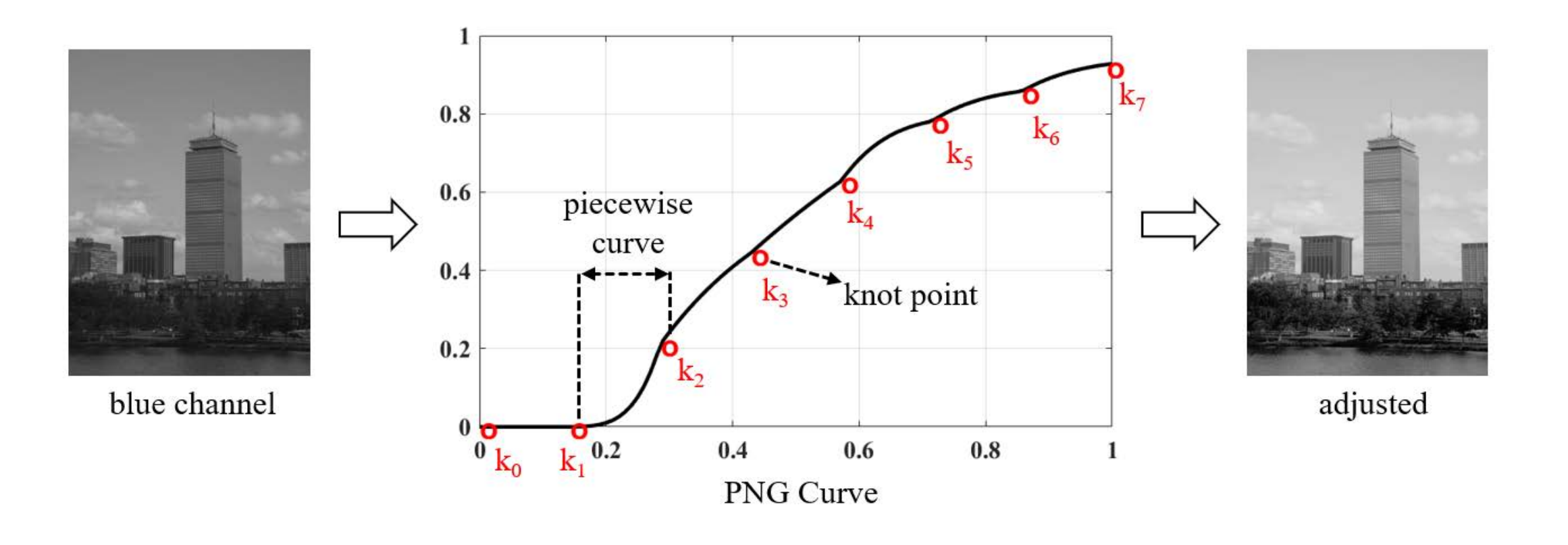}}
	\caption{An illustration of PNG Curve global adjustment. The black line represents a PNG Curve, wherein the red points represent the estimated knot points and the curve between two knot points is a piecewise curve. The horizontal axis and the vertical axis represent the input and output pixel values, respectively.}
	\label{fig:curves}
\end{figure*}

We separately apply one PNG Curve to each of the three RGB channels 
of a given image and yield one globally adjusted result. 
We further allow the network to produce multiple adjustment curves for curve mapping. This gives us multiple complementary solutions and each performs differently on bright and dark regions. Curve mapping via multiple sets of PNG Curves produces multiple globally adjusted results. We provide more discussion in Section~\ref{sec:SCM-Fusion}.

We typically have a small set of knot points and curve parameters for each PNG Curve, the estimation of which only requires a lightweight network. The effects of using different number of knot points and curve parameters will be investigated in the ablation study. 
In Figure \ref{fig:curves}, we show an example of adjusting the blue channel of an image using a PNG Curve. As shown, the PNG Curve adjusts the input pixel values in a piecewise manner. It nonlinearly increases or decreases the dynamic range of input pixels in each piece. Improved brightness and contrast are observed after such global adjustment.

\subsection{Multi-Task Network}
\label{sec:MT-Net}
The knot points, curve parameters, and confidence maps are closely related for obtaining the final result.
Consequently, we estimate them in a multi-task framework. We use a common backbone to extract shared features and fork different branches for the estimation of the aforementioned outputs.
We design the MT-Net to be lightweight -- it only consists of 153,488 trainable parameters. One can replace the proposed MT-Net with a more powerful and complex network at the cost of higher computations. As a baseline model, we only adopt a plain and parameter-efficient network structure.

\begin{figure}
	\centering
	\centerline{\includegraphics[width=0.5\textwidth,height=5cm]{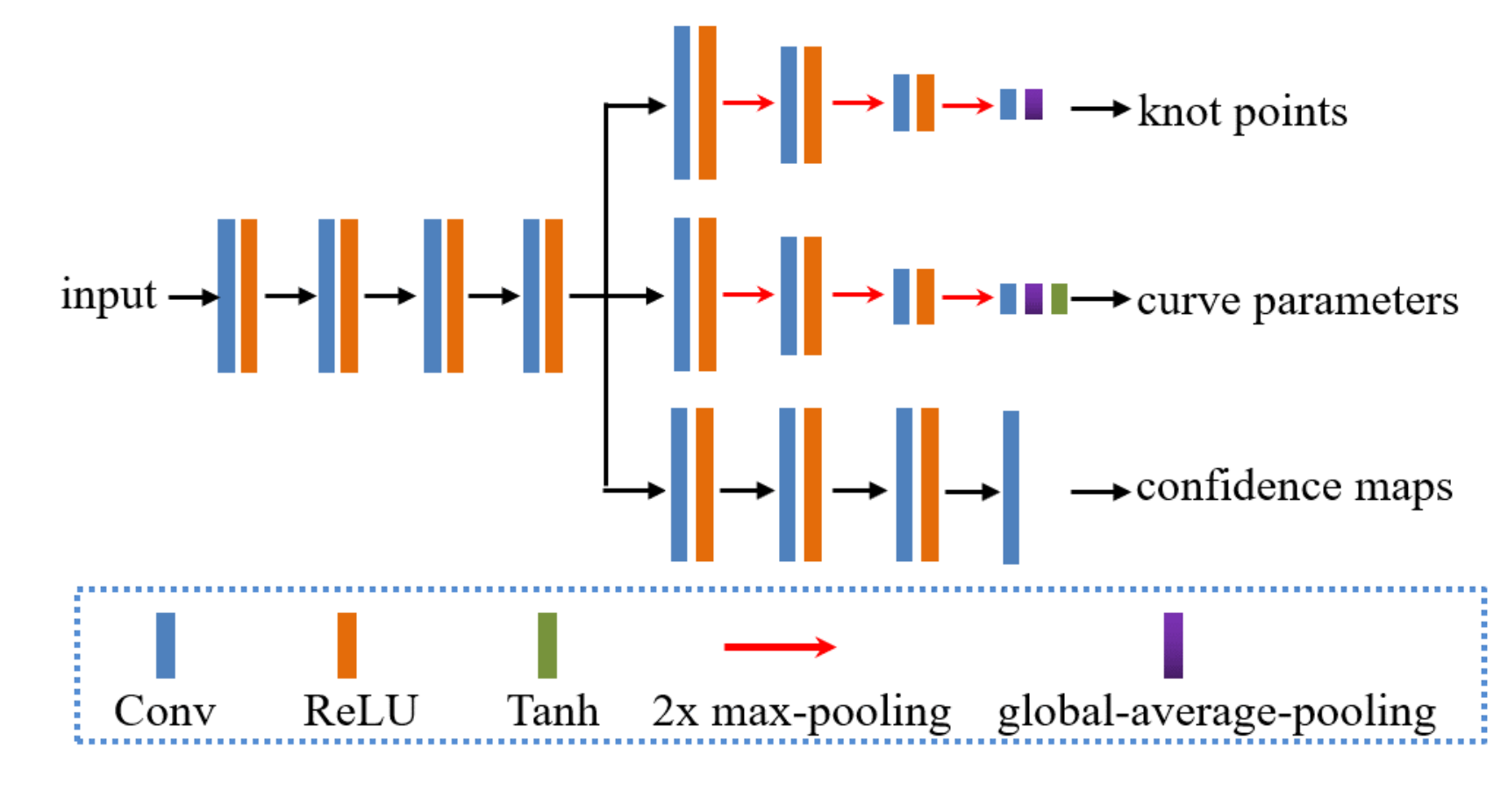}}
	\caption{The network structure of the proposed multi-task network (MT-Net).}
	\label{fig:MT-Net}
\end{figure}

As shown in Figure \ref{fig:MT-Net}, MT-Net only consists of 16 convolutional layers and each layer consists of 32 convolutional kernels of size 3$\times$3 and stride 1. 
Except for the last layer in each branch that outputs knot points, curve parameters, and confidence maps, respectively, all convolutional layers are followed by the ReLU activation function. 
Instead of the commonly used fully connected layers for parameter estimation in neural networks \cite{CURL,Li2018TIP1}, the knot points and curve parameters are obtained by global average pooling after the last convolutional layer, which reduces the computational resource costs and relaxes the sizes of input image. We constrain the curve parameter $\alpha$ of each piece in the range of [-1,1] by using the Tanh activation function. 
In both the knot point estimation branch and the curve parameter estimation branch, we employ the 2$\times$ max-pooling operations to enlarge the receptive field and reduce the computational burden.

In this work, we adopt eight knot points and seven sets of curve parameters for each PNG Curve. There is a PNG Curve for each of the three RGB channels. Here, we assume $N$ globally adjusted results.
Thus, the three branches of MT-Net produce $N$$\times$3$\times$8 knot points, $N$$\times$3$\times$7 sets of curve parameters, and $N$$\times$3 confidence maps, respectively.  

\subsection{Spatial-Adaptive Confidence Map Fusion}
\label{sec:SCM-Fusion}
Curve adjustment performs global enhancement and thus falls short in considering different characteristics of local regions. To avoid over-/under-enhancement in local regions, we formulate MT-Net so that it can produce multiple solutions. The different characteristics in different regions of multiple solutions are fused to achieve better global and local enhancement.
In our framework, there is an option to add constraints on the SCM Fusion so that each of global solutions can serve as a standalone and sensible output and thus allow spatial interpolation to achieve new results. This mechanism provides additional flexibility to explore different enhancement favors. Next, we present different SCM Fusion strategies, including the plain SCM Fusion, SCM Fusion constrained by separate training sets, and SCM Fusion with constraints on confidence maps. 

\begin{figure*}[t]
	\begin{center}
		\begin{tabular}{c@{ }c@{ }c@{ }c@{ }c}
			\includegraphics[width=.2\textwidth,height=2.6cm]{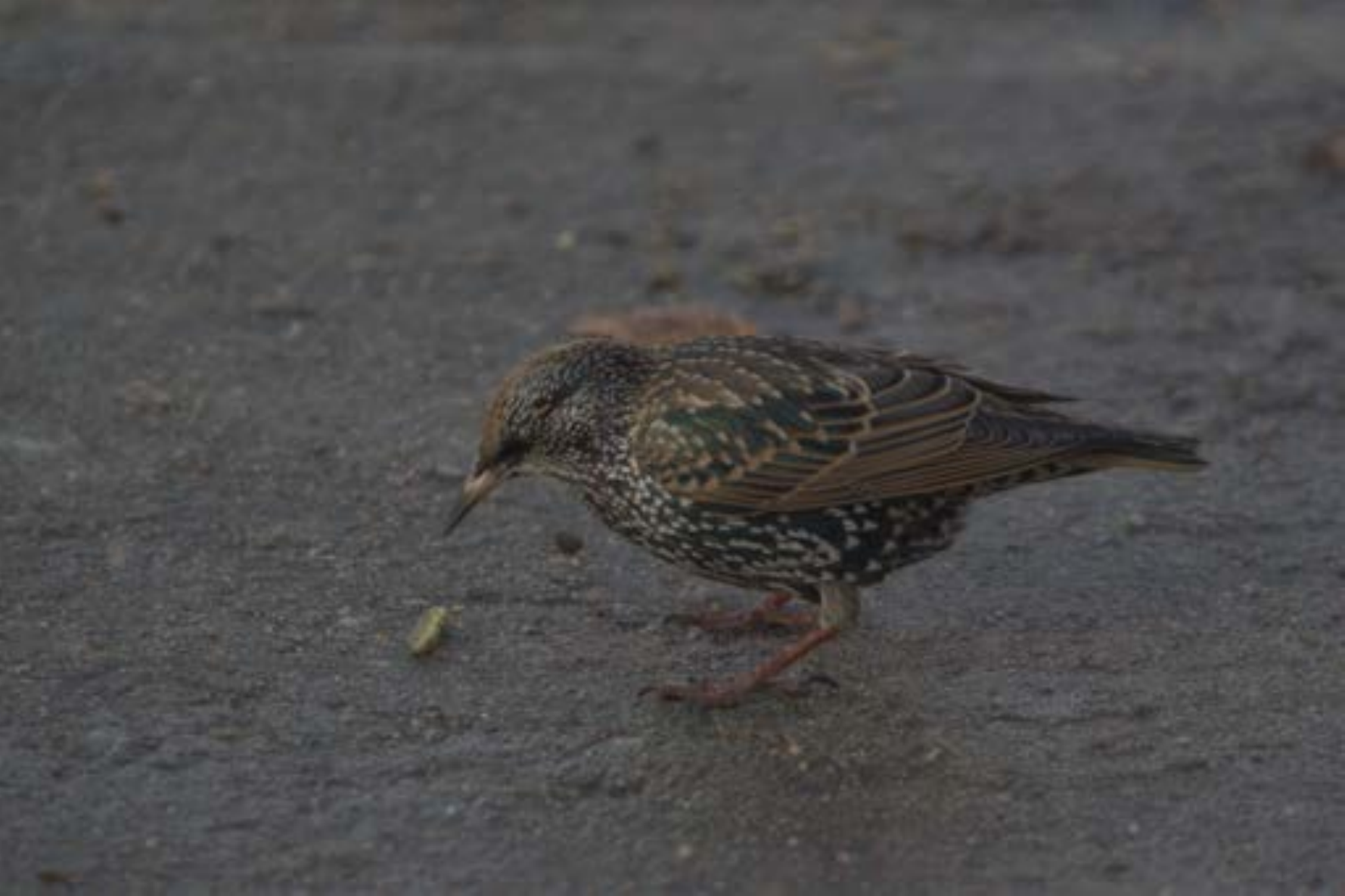}&
			\includegraphics[width=.2\textwidth,height=2.6cm]{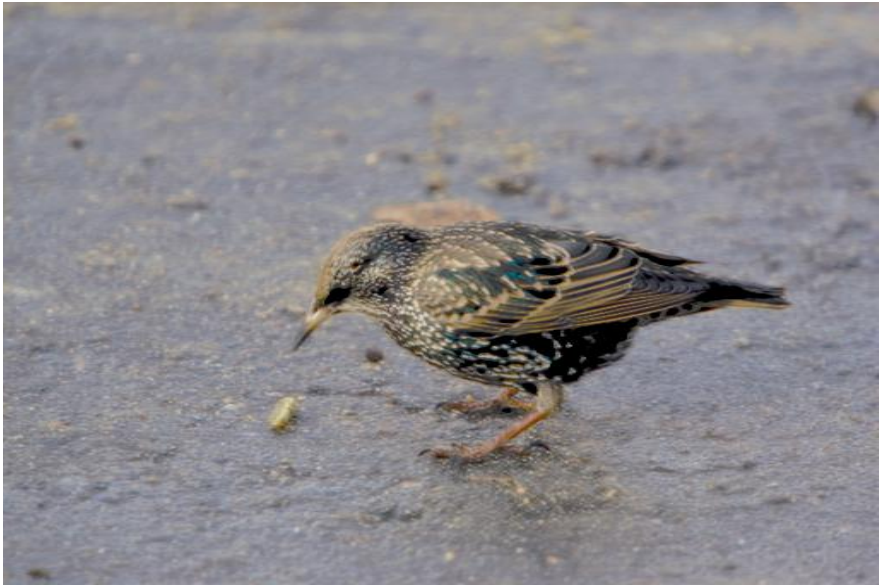}&
			\includegraphics[width=.2\textwidth,height=2.6cm]{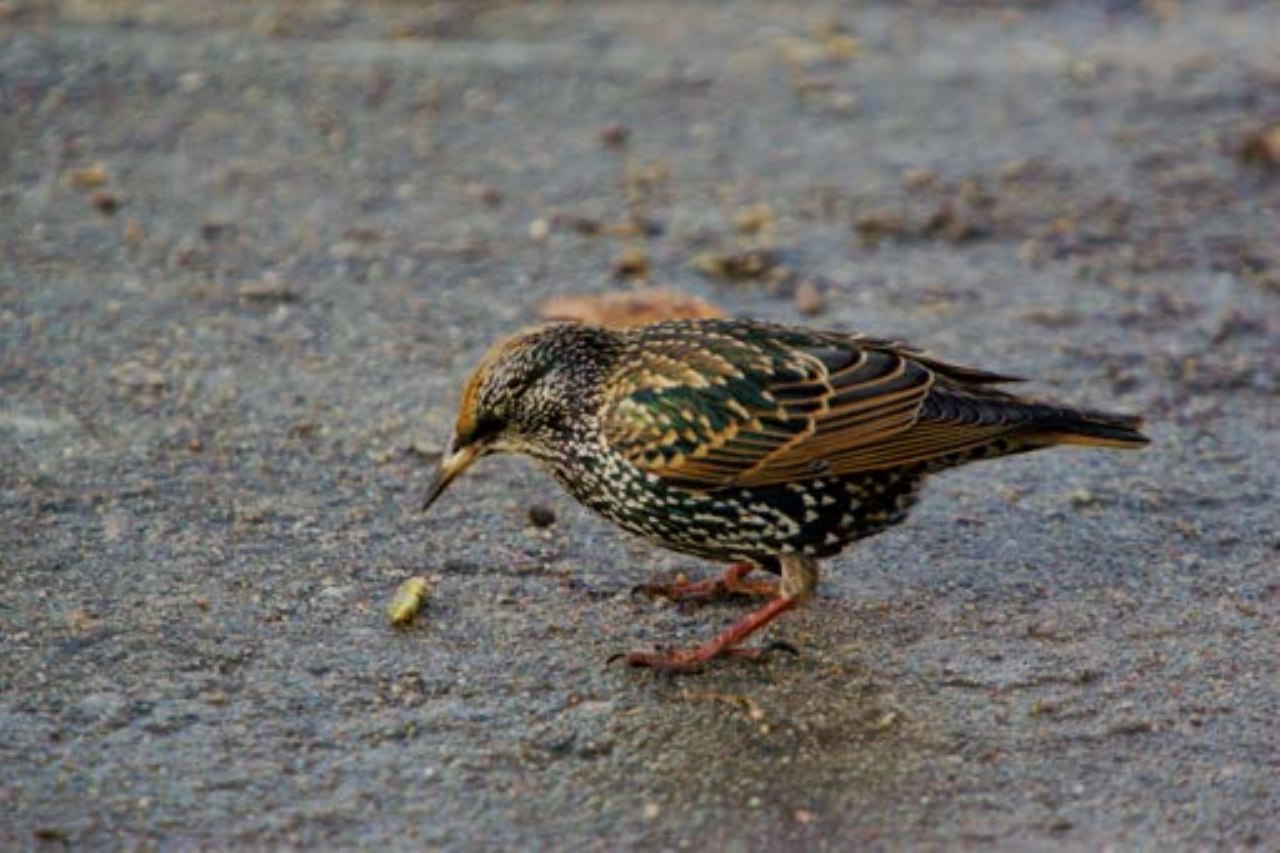}~&
			\includegraphics[width=.2\textwidth,height=2.6cm]{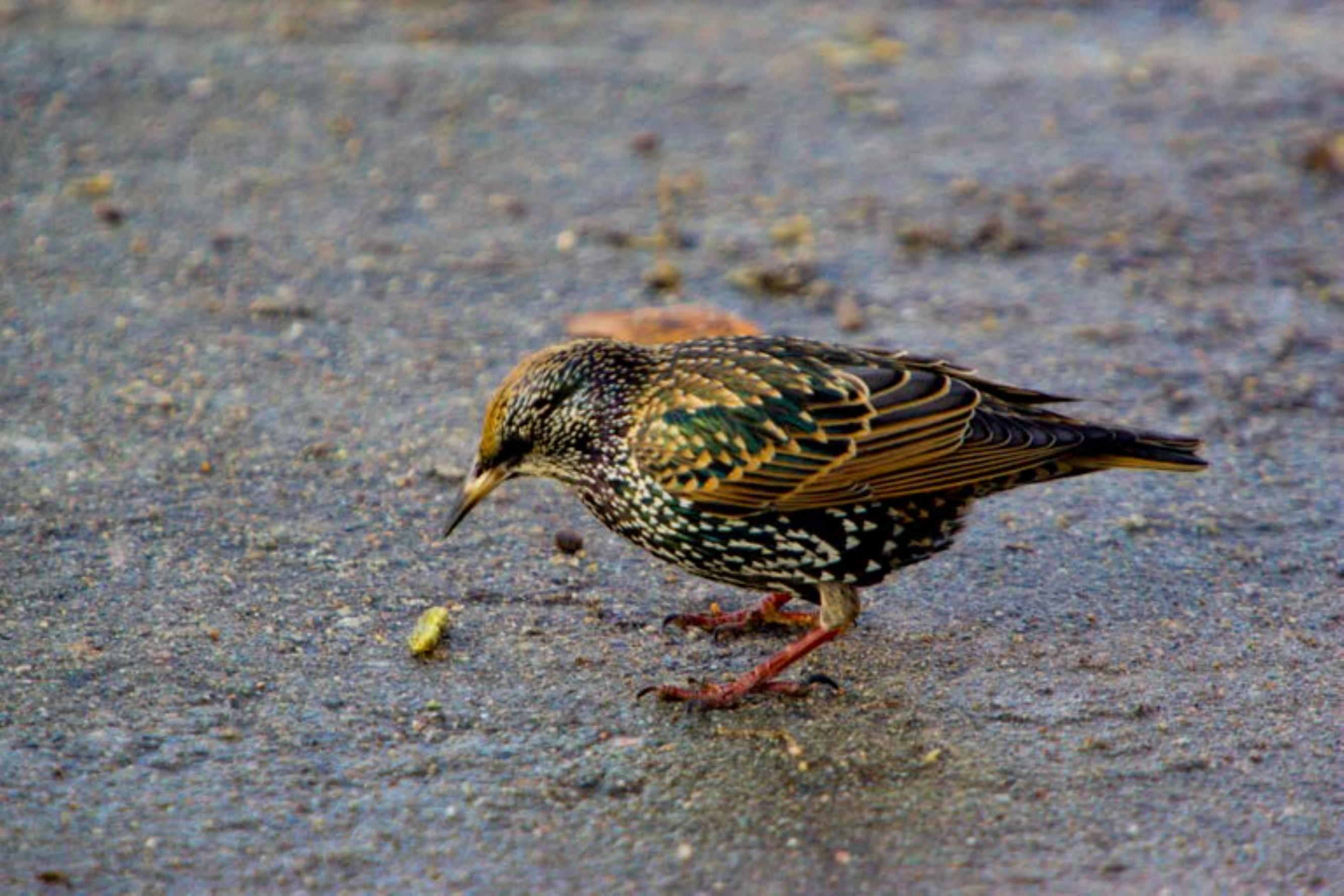}\\
			(a) input & (b) -w/o SCN Fusion & (c) -w plain SCN Fusion & (d) ground truth \\
		\end{tabular}
	\end{center}
	\caption{A visual comparison between FlexiCurve-pair-w/ plain SCN Fusion and FlexiCurve-pair-w/o SCN Fusion.}
	\label{fig:plain_fusion}
\end{figure*}

\subsubsection{Plain SCM Fusion}
\label{sec:Plain SCM Fusion}
To minimize the deviation between the final result and the ideal result (e.g., the ground truth), we design a plain SCM Fusion to fuse multiple globally adjusted results. Here we assume three globally adjusted outputs (each comprises three RGB channels). We found that fusing more global outputs does not improve performance further but raising the difficulty of stable training within the unpaired framework. 
The plain SCM Fusion can be expressed as:
\begin{equation}
\label{equ_5}
R=V_{1}\odot C_{1}\oplus V_{2}\odot C_{2}\oplus V_{3}\odot C_{3},
\end{equation}
where $R$ is the final result,  three globally adjusted results are represented by $V_{1}$, $V_{2}$, and $V_{3}$, respectively, the estimated confidence maps are $C_{1}$, $C_{2}$, and $C_{3}$, which have the same size as the input image (i.e., the same height, width, and channel), we denote the element-wise multiplication and element-wise addition as $\odot$ and $\oplus$, respectively.

\begin{figure*}[!t]
	\centering
	\centerline{\includegraphics[width=0.95\textwidth]{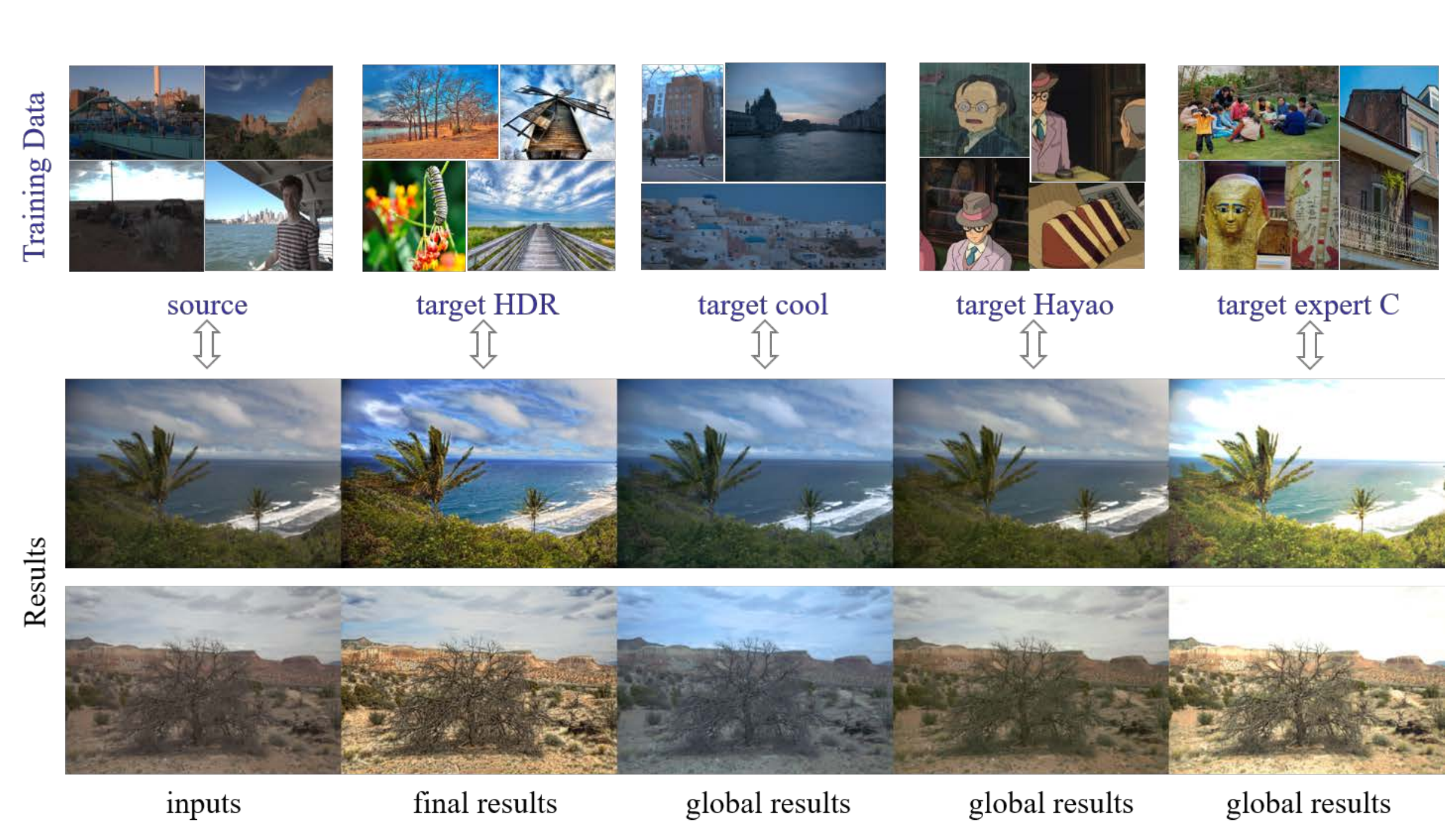}}
	\caption{Visual examples of SCN Fusion constrained by separate training sets. The top represents the sampled training data of source and target manifold. The bottom represents the final result and globally adjusted results by FlexiCurve-unpair with the SCM Fusion constrained by separate training sets including HDR, cool tone, Hayao, and expert C training sets (from left to right).}
	\label{fig:one2many}
\end{figure*}

The plain SCM Fusion effectively minimizes the deviation between the final result and the ideal result in comparison with just assuming one globally adjusted result. The reason can be shown as follows. Taking a color channel of an input RGB image as an example, the points $C$, $D$, and $E$ represent ideal output pixel values of three input pixels with identical input pixel value but at different spatial positions in this channel. Performing global adjustment would map all the identical input pixel value to a point $A$, the resulting deviation $\Delta_{2D}$ can be expressed as:
\begin{equation}
\label{equ_18}
\Delta_{2D}=\|CA\|+\|DA\|+\|EA\|,
\end{equation}
where $\|CA\|$, $\|DA\|$, and $\|EA\|$ represent the distance between the globally adjusted point $A$ and the ideal points $C$, $D$, and $E$, respectively. In the process of optimizing global curve parameters, the purpose is to find  a point $A$ that minimizes the deviation $\Delta_{2D}$.  Nevertheless, it is impossible for global  adjustment to reduce such deviation to zero.
In contrast, the plain SCM Fusion can  fuse several  global output pixel values to achieve the ideal output pixel value by the adaptive confidence maps that are optimized to minimize the deviation.  In Figure \ref{fig:plain_fusion}, we show a visual comparison between the plain SCM Fusion and a baseline model without SCM Fusion (i.e., we only produce one globally adjusted output and constrain it as the final result). As observed in Figure \ref{fig:plain_fusion}, since FlexiCurve with plain SCM Fusion can generate diverse global outputs, it is capable of covering broad solution spaces, which in turn reduces deviation against the ground truth. Although the baseline model FlexiCurve without SCN Fusion can improve the brightness of input image, it is still a global adjustment, which leads to the loss of local details in the result shown in Figure \ref{fig:plain_fusion}(b).

\subsubsection{SCM Fusion Constrained by Separate Training Sets}
\label{Separate Training Sets}
A globally adjusted output among multiple solutions is not meaningful on its own, because we separately enhance the three RGB channels by piecewise curves and do not constrain the relations of three RGB channels in a globally adjusted output in the training phase. 
While the current framework with plain SCM Fusion already suffices to produce a good enhancement result, one can further constrain the globally adjusted outputs by using separate training sets so that each of them can serve as a standalone and sensible globally adjusted result.

\begin{figure*}[t]
	\centering
	\centerline{\includegraphics[width=1\textwidth]{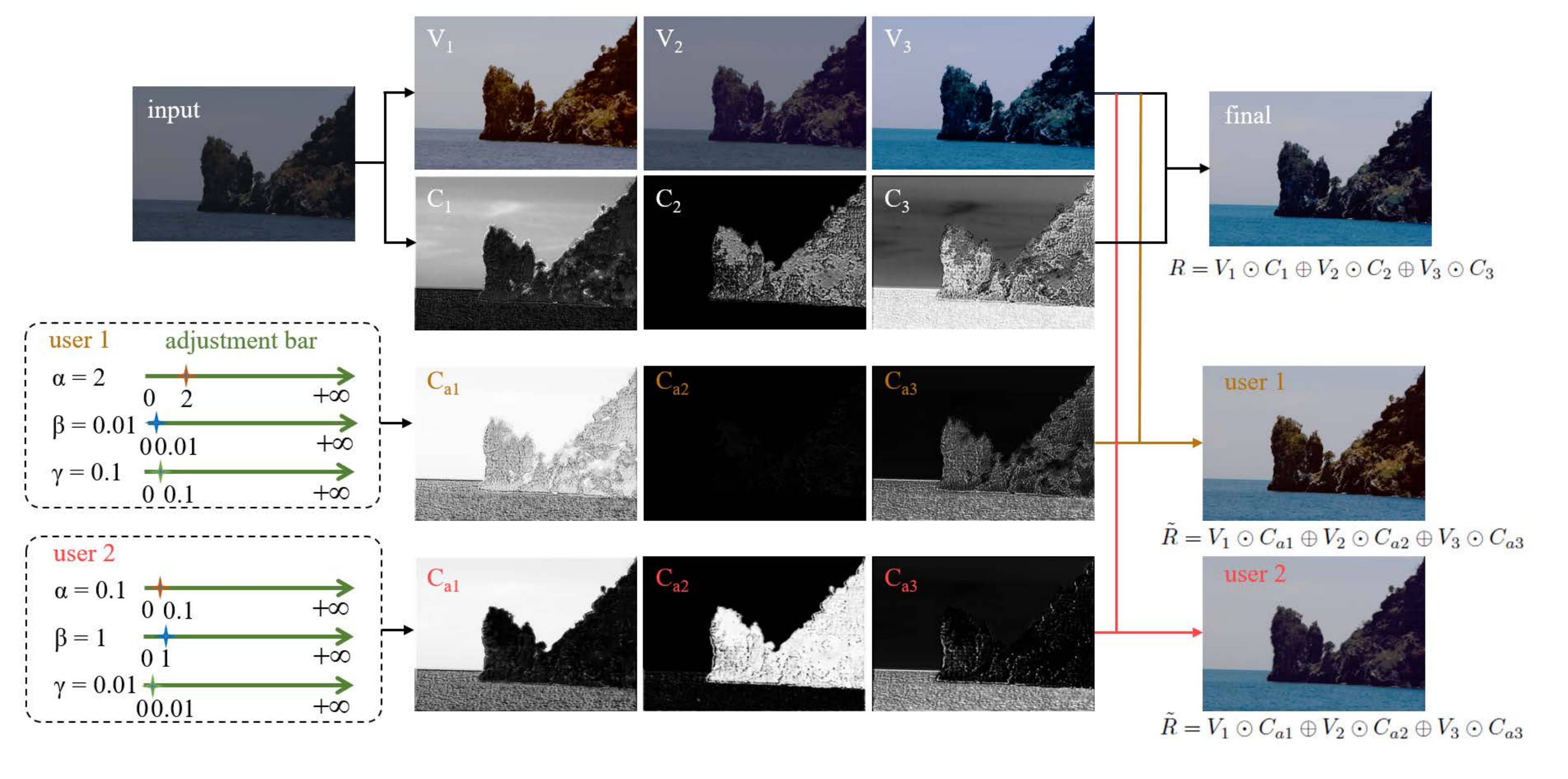}}
	\caption{Visual results of spatial interpolation by FlexiCurve-unpair with SCM Fusion with constraints on confidence maps. This FlexiCurve-unpair is constrained by separate training sets, including target expert C  for final result $R$, target Hayao, target cool, and target landscape for global adjusted results $V_{1}$, $V_{2}$, and $V_{3}$, respectively. The estimated confidence maps by MT-Net are represented as $C_{1}$, $C_{2}$, and $C_{3}$ while the adjusted confidence maps by different adjustment weights  $\alpha$, $\beta$, and $\gamma$ are represented as $C_{a1}$, $C_{a2}$, and $C_{a3}$, respectively. `final' represents the final result by plain SCM Fusion, indicated as $R$. `user' is the spatial interpolated results via simiple assigning larger adjustment weights to the prefer globally adjusted result, indicated as $\tilde{R}$. Here, the adjustment weights of $\alpha$, $\beta$, and $\gamma$ are set to 2, 0.01, and 0.1 as well as 0.1, 1, and 0.01 by user 1 and user 2, respectively.}
	\label{fig:user}
\end{figure*}

To be specific, we can separately constrain each globally adjusted result and final result by using different training sets. In Figure \ref{fig:one2many}, we give two sets of visual results of SCM Fusion constrained by separate training sets. As shown, the target HDR training set is associated with the final results, while the target cool, target Hayao, and target expert C training sets are used for the three globally adjusted results in the training phase, respectively. More experimental settings for generating Figure \ref{fig:one2many} can be found in Section \ref{Settings}. As shown in Figure \ref{fig:one2many}, the final results are similar to the HDR training set while the globally adjusted results are standalone and sensible and similar to the corresponding training sets. Nonetheless, the final results have better local details than globally adjusted results, thanks the fusion of multiple globally adjusted outputs. 
It is noteworthy that FlexiCurve with paired data can also produce standalone and sensible globally adjusted results when multiple paired training datasets are available.

\subsubsection{SCM Fusion with Constraints on Confidence Maps}

Since the standalone and sensible globally adjusted results have diverse properties such as different tone, brightness, contrast as shown in Figure \ref{fig:one2many}, the user may prefer one of the globally adjusted results more. Here, we offer a spatial interpolation scheme in our framework and discuss its effect on the solution space of final result. 

\noindent\textbf{Spatial Interpolation.} We can further impose the constraints on the fusion to allow user to interpolate new results toward the preferred globally adjusted result. Concretely, Eq. \eqref{equ_5}  can be modified as:
\begin{equation}
\label{equ_16}
\tilde{R}=V_{1}\odot C_{a1}\oplus V_{2}\odot  C_{a2}\oplus V_{3}\odot C_{a3},
\end{equation}
where $\tilde{R}$ is the interpolated result,  $C_{a1}$, $C_{a2}$, and $C_{a3}$ are the adjusted confidence maps by simple assigning different weights. The adjusted confidence map $C_{ai}$ can be achieved by:
\begin{equation}
\label{equ_17}
C_{ai}=\left\{
\begin{aligned}
C_{i}*\alpha/(C_{1}*\alpha+C_{2}*\beta+C_{3}*\gamma) &  & i=1 \\
C_{i}*\beta/(C_{1}*\alpha+C_{2}*\beta+C_{3}*\gamma) &  & i=2\\
C_{i}*\gamma/(C_{1}*\alpha+C_{2}*\beta+C_{3}*\gamma) &  & i=3
\end{aligned}
\right.
\end{equation}
where $\alpha$, $\beta$, and $\gamma$ represent the adjustment weights that correspond to estimated confidence maps $C_{1}$, $C_{2}$, and $C_{3}$, respectively.  To implement spatial interpolation, in addition to the constraints of separate training sets mentioned in Section \ref{Separate Training Sets}, we further constrain the confidence map $C$: 1) it should be in the range of [0,1]. This can be achieved by introducing a Sigmoid activation function to the the last layer of the confidence map estimation branch, and 2) it should have the same size as the input image, but only consisting of one channel instead of three channels used in Eq. \eqref{equ_5}. The last two constraints treat the three RGB channels of a globally adjusted result as a whole in the process of fusion, and thus allows weight adjustment among different confidence maps by simple sliding scales.

We present some examples of spatial interpolation by FlexiCurve-unpair in Figure \ref{fig:user}. More experimental settings for generating Figure \ref{fig:user} can be found in Section \ref{Settings}. Assuming that  user 1 prefers the global result $V_{1}$, thus a higher adjustment weight is assigned to $V_{1}$, i.e., $\alpha$=2, $\beta$=0.01, and $\gamma$=0.1 as shown in Fig. \ref{fig:user}. As shown, the interpolated result (denoted as user 1) combines the local characteristics of the final result and global result $V_{1}$, such as the color of the mountain of $V_{1}$ is smoothly transferred to the interpolated result. For another example, assuming that  user 2 prefers the global result $V_{2}$ and sets the adjustment weights to $\alpha$=0.1, $\beta$=1, and $\gamma$=0.01. Hence, the spatially  interpolated result (denoted as user 2) achieves the color of the mountain of $V_{2}$. More diverse image collections and paired training datasets can be used for training such a flexible framework.  In this way, our method can produce more personalized interpolation results.

\noindent\textbf{Adding Constraints During Training.}

To implement spatial interpolation, we 1) reduce channel  numbers of each confidence map from three to one, and 2) normalize each confidence map in the range of [0,1]. These two constraints may affect the accuracy of the final result. Next, we discuss their effect on the solution space of the final result.

\begin{figure}
	\begin{center}
		\begin{tabular}{c@{ }c@{ }}
			\includegraphics[height=4.6cm,width=4.3cm]{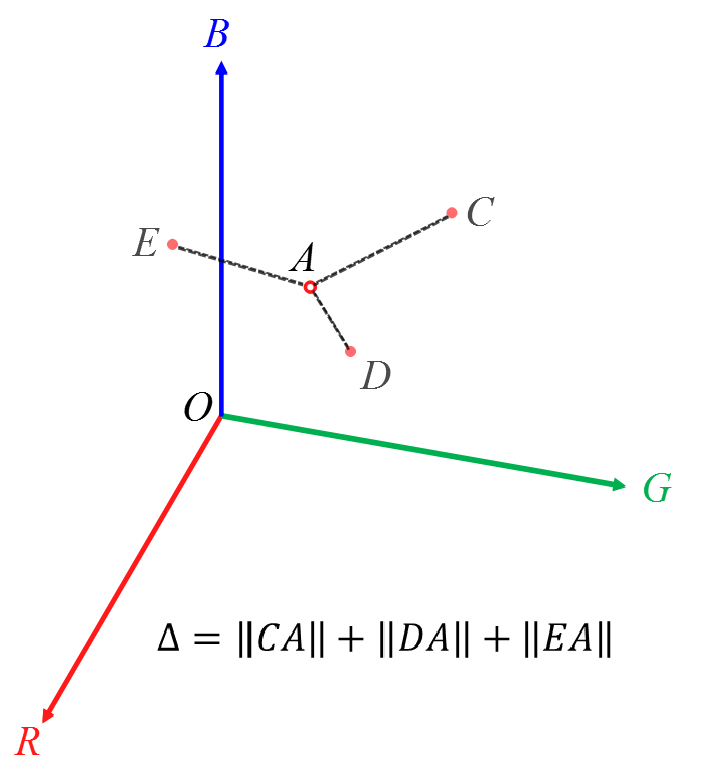}&
			\includegraphics[height=4.6cm,width=4.3cm]{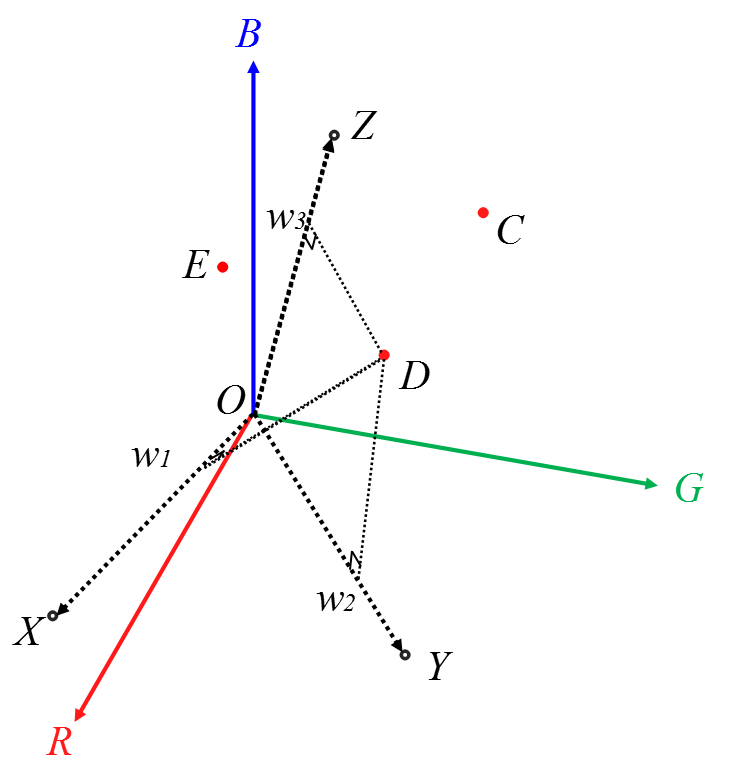} \\
			(a)&  (b)   \\
		\end{tabular}
	\end{center}
	\caption{An illustration of the potential deviation induced by (a) global adjustment and (b) the constraint of one channel confidence map.}
	\label{fig:global_curve_deviation}
\end{figure}

First, we constrain each confidence map to having only one channel. 
This extends the fusion process from 2D space to 3D space, i.e., from channel-wise fusion to image-wise fusion.
To analyze the deviation, an example of the deviation induced by global adjustment in RGB 3D space is shown in Figure \ref{fig:global_curve_deviation}(a). Same as the deviation induced by global adjustment in 2D space, the deviation 
$\Delta$ in RGB 3D space can be expressed as:
\begin{equation}
\label{equ_10_1}
\Delta=\|CA\|+\|DA\|+\|EA\|,
\end{equation}
where $\|CA\|$, $\|DA\|$, and $\|EA\|$ represent the distance between the globally adjusted point $\mathbf{A}$ and the ideal points $\mathbf{C}$, $\mathbf{D}$, and $\mathbf{E}$, respectively. Here, the points $\mathbf{A}$, $\mathbf{C}$, $\mathbf{D}$, and $\mathbf{E}$ are in the 3D space, which are different from the points $A$, $C$, $D$, and $E$ in 2D space mentioned in  Section \ref{sec:Plain SCM Fusion}.
In 3D space, a global adjustment  is also impossible to reduce such deviation $\Delta$  to zero.

The proposed SCM Fusion can help reduce the deviation $\Delta$. Assuming that we are fusing three globally adjusted results, and the three vectors between the center point $\mathbf{0}$ and the three globally adjusted  points $\mathbf{X}$, $\mathbf{Y}$, and $\mathbf{Z}$ are linearly independent, these three vectors can form a basis $\{$$\vec{X}$, $\vec{Y}$, $\vec{Z}$$\}$ as shown in Figure \ref{fig:global_curve_deviation}(b).
Theoretically, such a basis can constitute any pixel values in RGB 3D space, which makes the deviation $\Delta$ between an ideal pixel value and the constituted value being zero. The purpose of the network is to learn such a basis to minimize this deviation. For example, for an ideal point $\mathbf{D}$ in RGB 3D space shown in Figure \ref{fig:global_curve_deviation}(b), it can be constituted by:
\begin{equation}
\label{equ_20}
\mathbf{D}=w_{1}\vec{X}+w_{2}\vec{Y}+w_{3}\vec{Z},
\end{equation}
where $w_{1}$, $w_{2}$, and $w_{3}$ are the corresponding weights for  a basis $\{$$\vec{X}$, $\vec{Y}$, $\vec{Z}$$\}$, respectively. Hence, the constraint of one channel confidence map still has the potential of zero deviation.

\begin{figure}
	\centering
	\centerline{\includegraphics[width=0.35\textwidth,height=5.5cm]{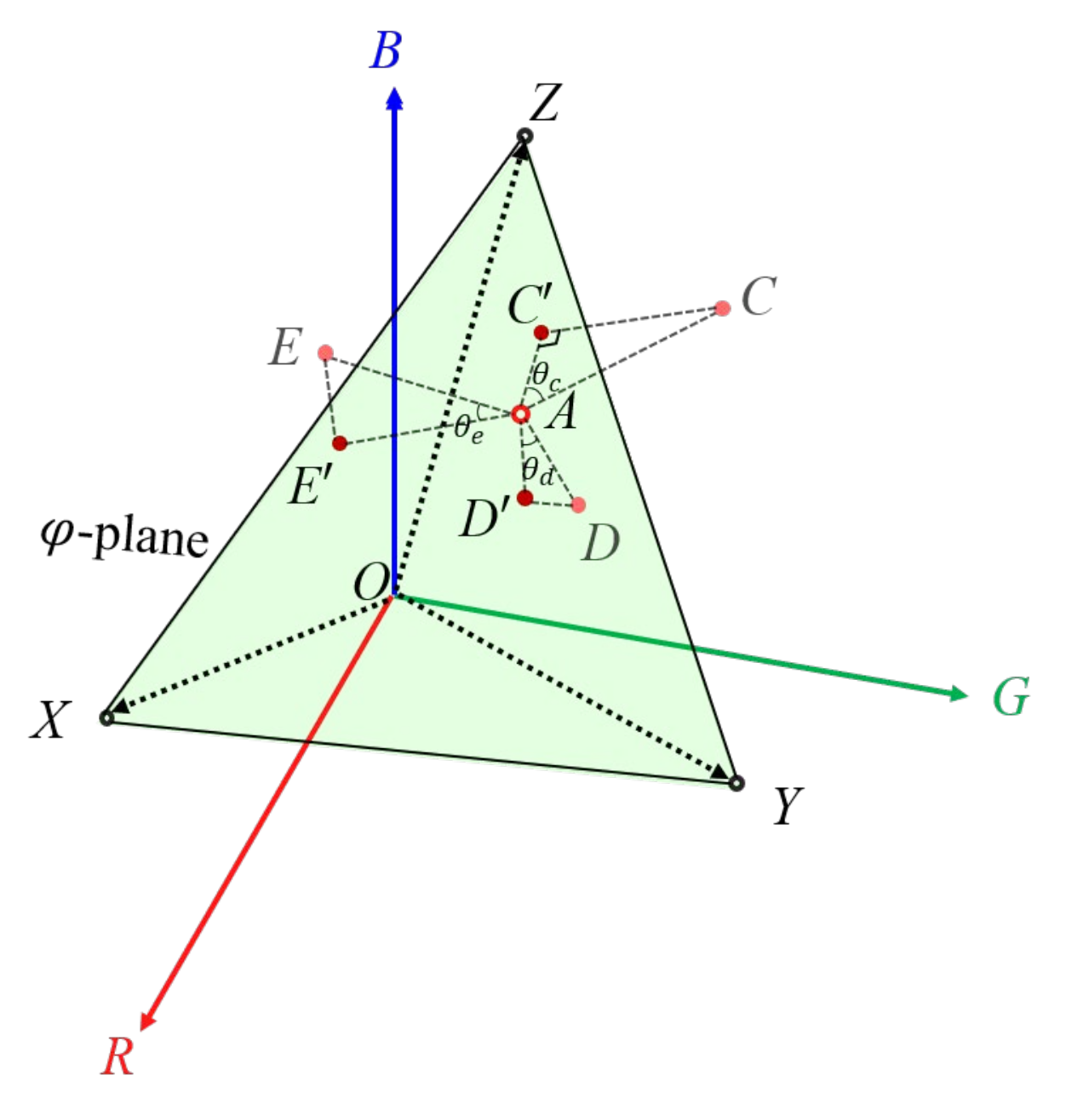}}
	\caption{An illustration of the potential deviation induced by the constraints of one channel normalized confidence map.}
	\label{fig:normalized_map_plane}
\end{figure}

Second, we further constrain the confidence maps normalized in the range of [0,1]. In this way, the whole RGB 3D space covered by the original basis $\{$$\vec{X}$, $\vec{Y}$, $\vec{Z}$$\}$ shown in  Figure \ref{fig:global_curve_deviation}(b) would be degraded to a smaller plane, such as the green plane $\varphi$ shown in Figure \ref{fig:normalized_map_plane} since the weights mentioned in Eq. \eqref{equ_20} are constrained in the range of [0,1]. 
We can find at least a plane  $\varphi$ that goes through the same point $\mathbf{A}$ as shown in Figure \ref{fig:global_curve_deviation}(a). For the ideal point values  $\mathbf{C}$, $\mathbf{D}$, and $\mathbf{E}$, the induced deviation by the basis $\{$$\vec{X}$, $\vec{Y}$, $\vec{Z}$$\}$ in this plane $\varphi$ with the constraints of one channel normalized confidence map can be expressed as:
\begin{equation}
\begin{aligned}
\label{equ_111}
\Delta_{\varphi}=&\|CC'\|+\|DD'\|+\|EE'\|\\
=&\|CA\|\sin\theta_{c}+\|DA\|\sin\theta_{d}+\|EA\|\sin\theta_{e},
\end{aligned}
\end{equation}
where $\mathbf{C'}$, $\mathbf{D'}$, and $\mathbf{E'}$ are the foot points of $\mathbf{C}$, $\mathbf{D}$, and $\mathbf{E}$ projected in the plane $\varphi$, respectively, and $\sin\theta_{c}$, $\sin\theta_{d}$, and $\sin\theta_{e}$ represent the  included angles between lines $\|CA\|$, $\|DA\|$, $\|EA\|$ and the plane $\varphi$, respectively. Consequently, $|CC'\|+\|DD'\|+\|EE'\|$ can be expressed as $\|CA\|\sin\theta_{c}+\|DA\|\sin\theta_{d}+\|EA\|\sin\theta_{e}$.
Compared with the deviation $\Delta$ in Eq. \eqref{equ_10_1}, Eq. \eqref{equ_111} can have 0 $\leq$ $\Delta_{\varphi}$ $\leq$  $\Delta$  because of 0 $\leq \sin\theta \leq$ 1. Such a result indicates that the deviation caused by the one channel normalized confidence map becomes larger than the original basis  $\{$$\vec{X}$, $\vec{Y}$, $\vec{Z}$$\}$ without any constraints (i.e., the deviation approximating zero) but smaller than the deviation induced by the global adjustment (i.e., the deviation $\Delta$ in Eq\eqref{equ_10_1}.

In summary, we achieve spatial interpolation with the constraints of one channel normalized confidence map at the cost of diverging from the reference collection of the final result (i.e., inducing larger deviation between the final result and the ideal result). If not specified, we adopt the plain SCM Fusion for more accurate final results.

\subsection{Loss Functions}
\label{sec:Loss Function}
We adopt different loss functions to optimize FlexiCurve with paired or unpaired data.

\subsubsection{Training with Unpaired Data}
We adopt adversarial training to train our model with unpaired data. 
Specifically, we treat MT-Net as a generator $G$, of which the goal is to produce outputs that fool the discriminators.
We have two discriminators i) a global discriminator $D_g$ that determines whether the image is from the target manifold, i.e., a reference image collection, and ii) a local discriminator $D_l$ that learns to distinguish between random local patches from the generator's output and arbitrary sampled image from the target manifold.
We iteratively update the generator and the discriminators in the training phase.  The total loss of $G$ can be expressed as:
\begin{equation}
\label{equ_6}
\begin{aligned}
L^{total}_{unpair}=W_{g}L_{G_{g}}+W_{l}L_{G_{l}}+L_{c},
\end{aligned}
\end{equation}
where $W_{g}$ and $W_{l}$ are the weights of the global adversarial loss $L_{G_{g}}$ and local adversarial loss $L_{G_{l}}$, respectively. The content loss is represented as  $L_{c}$. The losses are applied to the output after SCM Fusion. We introduce the content loss and adversarial losses as follows.

\noindent
\textbf{Content Loss.} To preserve the image semantic content, we adopt the perceptual loss \cite{perceptual} to constrain the perceptual similarity between the input image and the final result.
Specifically, we use the VGG-16 network \cite{VGG} pre-trained on ImageNet \cite{ImageNet} to extract high-level feature maps from the input image and the final result, and compute the feature space distance: 
\begin{equation}
\label{equ_7}
L_{c}=\frac{1}{HW}\sum_{i=1}^{H}\sum_{j=1}^{W} |\phi_{l}(R(i,j))-\phi_{l}(I(i,j))|,
\end{equation}
where $\phi_{l}(\cdot)$ denotes the $l$th convolutional layer of the VGG-16 network. Following previous work \cite{CartoonGAN}, we employ the  features extracted by `conv4\_4' layer in our implementation for stable training. $W$ and $H$ denote the width and height of feature maps, respectively. $R$ is the final result while $I$ represents the input image. 

\noindent
\textbf{Global Adversarial Loss.}
To encourage the final result to be indistinguishable from the images in the target manifold, we solve the following optimization problem:
\begin{equation}
\label{equ_8}
\min_{G}\max_{D_g}\mathbb{E}_{I\sim S}[log(1-D_g(G(I)))]+\mathbb{E}_{J\sim T}[log(D_g(J))],
\end{equation}
where the generator $G$ tries to generate `fake' images to fool the global discriminator $D_g$. The global discriminator $D_g$ tries to distinguish `fake' images from the reference images. The input $I$ is sampled from the source manifold $S$ while $J$ is the reference image arbitrary sampled from the target manifold $T$. For the generator, we minimize the following loss function:
\begin{equation}
\label{equ_9}
L_{G_{g}}=log(1-D_g(G(I))),
\end{equation}
where $G(I)$ produces final result. For the global discriminator, we minimize the loss function:
\begin{equation}
\label{equ_10}
L_{D_{g}}=-log(1-D_g(G(I)))-log(D_g(J)).
\end{equation}

\noindent
\textbf{Local Adversarial Loss.}
Similarly with the global adversarial loss, the local adversarial loss for updating the generator can be expressed as:
\begin{equation}
\label{equ_11}
L_{G_{l}}=\frac{1}{P}\sum_{p=1}^{P}log(1-D_l(G(I)^{p}_{patch})),
\end{equation}
where $P$ is the number of image patches, which is set to 5 in our implementations. The $p$th image patch with a size of 32$\times$32 randomly cropped from the final result is denoted as $G(I)^{p}_{patch}$. The local discriminator is not sensitive to the numbers and sizes of image patches. For updating the local discriminator, we minimize the following loss function:
\begin{equation}
\label{equ_12}
L_{D_{l}}=-log(1-D_g(G(I)^{p}_{patch}))-log(D_g(J^{p}_{patch})),
\end{equation}
where $J^{p}_{patch}$ denotes the $p$th  randomly cropped image patch from the reference image $J$.

\subsubsection{Training with Paired Data}
When paired data is available for training,
we use $\ell_{2}$  loss and SSIM loss to train our model. The total loss can be expressed as:  
\begin{equation}
\label{equ_13}
\begin{aligned}
L^{total}_{pair}=\ell_{2}+W_{s}L_{s},
\end{aligned}
\end{equation}
where $\ell_{2}$  and $L_{s}$ are the $\ell_{2}$ and SSIM losses, respectively, the weight of SSIM loss is represented as $W_{s}$. We introduce these two losses as follows.

\noindent
\textbf{$\ell_{2}$  Loss.} 
The $\ell_{2}$ loss measures the difference between the final  result $R$ and corresponding ground truth $J$ as:
\begin{equation}
\label{equ_14}
L_{\ell_{2}}=\frac{1}{PQ}\sum_{i=1}^{P}\sum_{j=1}^{Q}  (R(i,j)-J(i,j))^2,
\end{equation}
where $P$ and $Q$ denote the width and height of input image, respectively.

\noindent
\textbf{SSIM Loss.} We introduce the SSIM loss to impose the structure and texture similarity on the latent image, which can be expressed as:
\begin{equation}
\label{equ_15}
L_{s} = 1-\frac{1}{PQ}\sum_{i=1}^{P}\sum_{j=1}^{Q}  SSIM(R(i,j), J(i,j)).
\end{equation}
Based on the default setting in SSIM loss \cite{Zhao2017}, we set the image patch size, parameters $c_{1}$ and $c_{2}$ to 13$\times$13, 0.02, and 0.03, respectively.

\section{Experimental Results}
\label{sec:Experimental_Results}

\subsection{Experimental Settings}
\label{Settings}
\subsubsection{Training Sets}
\label{Training Set}
For a fair comparison, we use the same training and testing data as \cite{Enhancer}.  To be specific, we use the MIT-Adobe 5K dataset \cite{Adobe5K} that contains 5,000 images and the corresponding reference images retouched by five experts. Following \cite{Enhancer}, we split the dataset into three partitions: 2,250 images and their retouched versions are used for paired training, the retouched versions of another 2,250 images are treated as the reference results for unpaired training, and the rest 500 images are used for testing. Moreover, the retouching results by experts C are treated as the reference images. In the training data, the input images and the reference images are exclusive. All images in the MIT-Adobe 5K dataset are decoded into the png format and resized to have a long-edge of 512 pixels by using Lightroom as the processing in \cite{Enhancer}. However, our method, in fact, can process any sizes of images.

To verify the performance of SCN Fusion constrained by separate training sets and spatial interpolation, we collect more training datasets to train FlexiCurve with unpaired data, including one source image dataset and five target image datasets. Specifically, we randomly sample images from the low-quality images of Adobe 5K dataset \cite{Adobe5K} as the source image dataset. We collect HDR images from Flickr (denoted as target HDR) and landscape images from Google Image (denoted as target landscape).
We use the Hayao cartoon style images sampled from \cite{CartoonGAN} (denoted as target Hayao), randomly sample images from retouched results by expert C from \cite{Adobe5K} (denoted as target expert C), and generate cool tone images by using Photoshop to process the images in the source image dataset (denoted as target cool).  Each dataset contains 600 images saved as png format for training and each image is randomly cropped to a size of 256$\times$256 in the training phase.

\subsubsection{Testing Sets}
\label{Testing Set}
We denote the 500 pairs of testing data in MIT-Adobe 5K dataset \cite{Adobe5K} as Adobe 5K-T in this paper.  In addition to the Adobe 5K-T testing  set, we also conduct experiments on extra testing data, i.e., CUHK-PhotoQuality Dataset (CUHK-PQ) \cite{CUHKPQ}. We randomly select ten low-quality images from each subset (total of eight subsets including animal, architecture, human, etc.) and thus form 80 low-quality testing images, denoted as CUHK-PQ testing set. We also show the results of applying our method to enhance the oldest color photos in the world collected from the Internet \cite{oldphoto}. 

\subsubsection{Implementation Details}
\label{Implementation Details}
We implement our framework with PyTorch on an NVIDIA Tesla V100 GPU.
A batch size of $20$ is applied. The filter weights of each layer are initialized with standard zero mean and 0.02 standard deviation Gaussian function. Bias is initialized as a constant. We use ADAM \cite{ADAM} optimizer with default parameters for optimizing our frameworks. 

For FlexiCurve with unpaired data (denoted as \textbf{FlexiCurve-unpair}), we first train the generator with  the content loss to improve the stability of training. After 15 epochs, we introduce adversarial losses and iteratively optimize the generator and discriminators until convergence. The global and local discriminators adopt the same network structure as the discriminator employed in \cite{CartoonGAN} based on its stability and efficiency. We use the fixed learning rate $5e^{-5}$ for our generator and discriminators optimization.
Both weights $W_{g}$ and  $W_{l}$ are set to 6 to balance the scale of losses. 
For FlexiCurve with paired data (denoted as \textbf{FlexiCurve-pair}), we set the initial learning rate to $1e^{-4}$ and then decreased  it to $1e^{-5}$ after 100 epochs  until convergence. We set the weight $W_{s}$ to 0.1. 

\subsubsection{Baselines}
\label{Comparison Methods}
We qualitatively and quantitatively compare our method with Contrast Limited Adaptive Histogram Equalization (CLAHE) \cite{CLAHE}, Adobe PhotoShop Auto (denoted as PS Auto), FCN \cite{Chen17}, DPE \cite{Enhancer}, DeepLPF \cite{Moran20}, DPED \cite{DSLR}, Exposure \cite{WhiteBox}, EnlightenGAN \cite{Jiang2019}, and CycleGAN \cite{CycleGANs}. 

For PS Auto, we select Ajustment$\rightarrow$Curves$\rightarrow$Options$\rightarrow$
Enhance Brightness and Contrast in Photoshop software. 
We directly use the FCN \cite{Chen17} model that approaches  the multiscale tone manipulation.  
We adopt the supervised DPE model \cite{Enhancer} trained by the same training data as our FlexiCurve-pair. We found that the results reported in DPE \cite{Enhancer} cannot be reproduced.  The same findings are also mentioned in recent work \cite{Kneubuehler}. For a fair comparison, we still use the original PSNR and SSIM values reported in their paper. However, we discard their weakly supervised model since its original PSNR and SSIM values are not reported in their paper. Instead, we retrain the CycleGAN as a baseline by using the same unpaired training data as our  FlexiCurve-unpair.   
For DPED, we directly use the models trained by the paired iPhone 3GS and Digital Single-Lens Reflex (DSLR) data and  Sony Xperia Z and  DSLR data, denoted as DPED\_Iphone and DPED\_Sony, respectively. For Exposure \cite{WhiteBox} and EnlightenGAN \cite{Jiang2019}, we directly use their released codes.
For DeepLPF \cite{Moran20}, the pre-trained models and results are unavailable. Thus, we only report the quantitative results in their paper.

\subsection{Ablation Study}

We conduct a series of ablation studies to investigate the effects of network hyperparameters, the effects  of curve parameters, the advantages of spatial-adaptive fusion, the effects of input image sizes, the benefits of spatial interpolation, and the contributions of local discriminator. 

We retrain the ablated models while keeping the same settings as our final model (i.e., FlexiCurve), except for the ablated parts. Quantitative comparisons in terms of full-reference image quality assessment (IQA) metrics Peak Signal-to-Noise Ratio (PSNR,dB) and Structural Similarity (SSIM)~\cite{SSIM} on Adobe 5K-T testing set are presented in Table \ref{table1}. We quantitatively study the effects of network hyperparameters, curve parameters, and spatial-adaptive fusion on the FlexiCurve-pair framework based on its stable training. We did not present the visual comparisons for the ablated models of network hyperparameters and curve parameters due to the subtle visual differences.

\begin{table}[!t]
	\centering
	\caption{Quantitative comparisons of the ablated models that refer to network hyperparameters (MT-Net), curve parameters (PNG Curve), spatial-adaptive fusion (SCM Fusion). The final FlexiCurve is equiped with 32 output feature maps in each convolutional layer denoted as $f$32, 16 convulutional layers denoted as $l$16,  7 knot points in each PNG  Curve denoted as $k$7, 4 iterations of the nonlinear adjustment curve in each piece denoted as $i$4, 3 globally adjusted results denoted as $N$3, i.e., FlexiCurve-$f$32-$l$16-$k$7-$i$4-$N$3.}
	\begin{center}
		\begin{tabular}{c|c|c|c}
			\hline
			\textbf{Ablated Models} & \textbf{Baselines} & \textbf{PSNR$\uparrow$} & \textbf{SSIM$\uparrow$}  \\
			\hline
			&	input image               &  17.82   & 0.78       \\
			&\textbf{FlexiCurve-pair}& \textbf{23.97}   & \textbf{0.91}  \\
			&($f$32-$l$16-$k$7-$i$4-$N$3)&    &   \\
			\hline
			\multirow{5}{*}{MT-Net}& $f$8-$l$16    &  21.83     &   0.87    \\
			&	$f$16-$l$16   &  22.95    & 0.88       \\
			&	$f$64-$l$16   &  23.99    & 0.91      \\
			&	$f$32-$l$20   &  24.00    & 0.90      \\
			&   $f$32-$l$24   &  24.03    & 0.90       \\
			\hline
			\multirow{5}{*}{PNG Curve}&		$k$7-$i$1    &    22.26   & 0.88      \\
			&	$k$7-$i$6     &   23.95    &  0.90     \\
			&	$k$10-$i$4    &   24.01    &  0.90     \\
			&	$k$16-$i$4    &   23.65    &  0.89     \\
			&	$k$32-$i$4    &   23.25    &  0.89     \\
			\hline
			\multirow{4}{*}{SCM Fusion}&$N$1 (global) & 22.01 &  0.89\\
			&   $N$2   &  23.52     &  0.90     \\
			&	$N$4   &  23.93     &  0.91     \\
			&	$N$5   &  23.91     &  0.91     \\
			\hline
		\end{tabular}
	\end{center}
	\label{table1}
\end{table}

\begin{table}[!t]
	\centering
	\caption{Quantitative comparisons of the FlexiCurve-pair with different sizes of input image in terms of PSNR (in dB), SSIM, and FLOPs (in G). We assume the original resolution of an input image is 512$\times$512$\times$3.}
	\begin{center}
		\begin{tabular}{c|c|c|c}
			\hline
			\textbf{Input Sizes}  & \textbf{PSNR$\uparrow$} & \textbf{SSIM$\uparrow$} & \textbf{FLOPs$\downarrow$} \\
			\hline
			original resolution           & 23.97   & 0.91 &    21.97   \\
			2$\times$ downsampling      &  23.26   & 0.90 &    5.49   \\
			3$\times$ downsampling      &  21.93   & 0.87  &   2.38  \\
			4$\times$ downsampling      &  20.47   & 0.83 &    1.37   \\
			\hline
		\end{tabular}
	\end{center}
	\label{table_inputsize}
\end{table}

\noindent
\textbf{Effect of Network Hyperparameters.}
We first study the effects of network hyperparameter settings in our MT-Net, consisting of the network's depth and width. In the ablated models of MT-Net, $f$ represents the number of output feature maps in each convolutional layer (except for the last layer in each branch) and $l$ represents the number of convolutional layers, where the feature extractor and three branches have the same numbers of convolutional layers (i.e., $l$/4).

As presented in Table \ref{table1}, increasing more feature maps $f$ from 8 to 32 (our final choice) improves the quantitative performance while increasing from 32 to 64 only slightly improves the performance at the cost of more computational resources. When increasing the number of convolutional layers $l$ from 16 to 24, we did not find obvious gains.  
Thus, we adopt $f$32 and $l$16 for MT-Net.

\noindent
\textbf{Effect of Curve Parameters.} We investigate  the effects of curve parameter settings in the PNG Curve, including the number of knot points (also related to the number of pieces) and iterations employed in the nonlinear adjustment curve in each piece. In the ablated models of PNG Curve, $k$ represents the number of knot points in each PNG Curve and $i$ represents the number of iterations of the nonlinear adjustment curve in each piece. 

Observing Table \ref{table1}, increasing the number of iterations from 1 to 4 significantly improves the values of PSNR and SSIM, suggesting the importance of flexible adjustment of curvature. However, further increasing the number of iterations from 4 to 6 cannot achieve obvious improvements because 4 iterations can handle the  most challenging cases.  Increasing the numbers of knot points, on the contrary, decreases the quantitative performance. The potential reason is that estimating more knot points increases the pressure of MT-Net under the limited network hyperparameters. Considering the minor differences between these ablated models, we adopt the settings of $k$7 and $i$4 for PNG Curve to balance the computational resources and enhancement performance.

\noindent
\textbf{Advantage of Spatial-Adaptive Fusion.}
To demonstrate the advantages of multiple globally adjusted results spatial-adaptive fusion, we provide the results of removing the spatial-adaptive fusion, i.e., only using the global piecewise curve to enhance input image, denoted as $N$1. As shown in Table \ref{table1}, single global enhancement shows poorer values of PSNR and SSIM when compared with the spatial-adaptive fusion results.
A visual comparison is shown in Figure \ref{fig:plain_fusion}. Both Table \ref{table1} and Figure \ref{fig:plain_fusion} validate our motivations of fusing multiple globally adjusted results instead of only using single global piecewise curve. However, we found that increasing the number of globally adjusted results from 3 to 5 slightly decreases the quantitative performance. This is because producing more globally adjusted results in our framework requires more piecewise curves and confidence maps, which is beyond our current network capacity.
Thus, we set the number of globally adjusted results to 3.

\begin{figure}[t]
	\begin{center}
		\begin{tabular}{c@{ }c@{ }c@{ }c}
			\includegraphics[width=.15\textwidth,height=5cm]{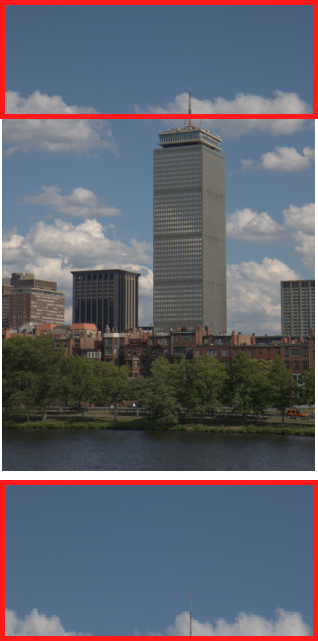}&
			\includegraphics[width=.15\textwidth,height=5cm]{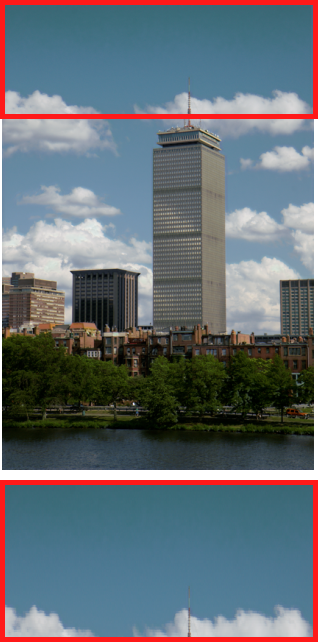}&
			\includegraphics[width=.15\textwidth,height=5cm]{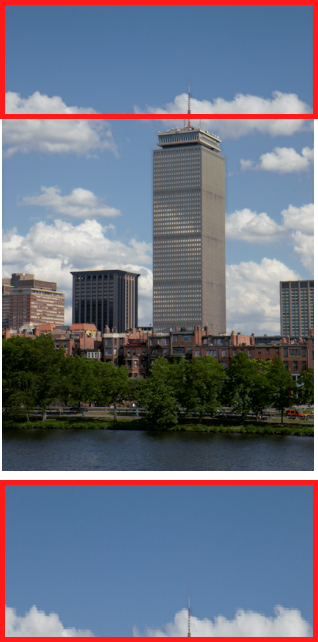}\\
			(a) input image & (b) automatic & (c) interpolated \\
		\end{tabular}
	\end{center}
	\caption{A visual comparison between  automatically produced result by FlexiCurve-unpair and spatially interplolated result ($\alpha$=1, $\beta$=0.1, $\gamma$=3). Red boxes indicate the amplified details.}
	\label{fig:si}
\end{figure}

\begin{figure}[t]
	\begin{center}
		\begin{tabular}{c@{ }c@{ }c@{ }c}
			\includegraphics[width=.15\textwidth,height=3.5cm]{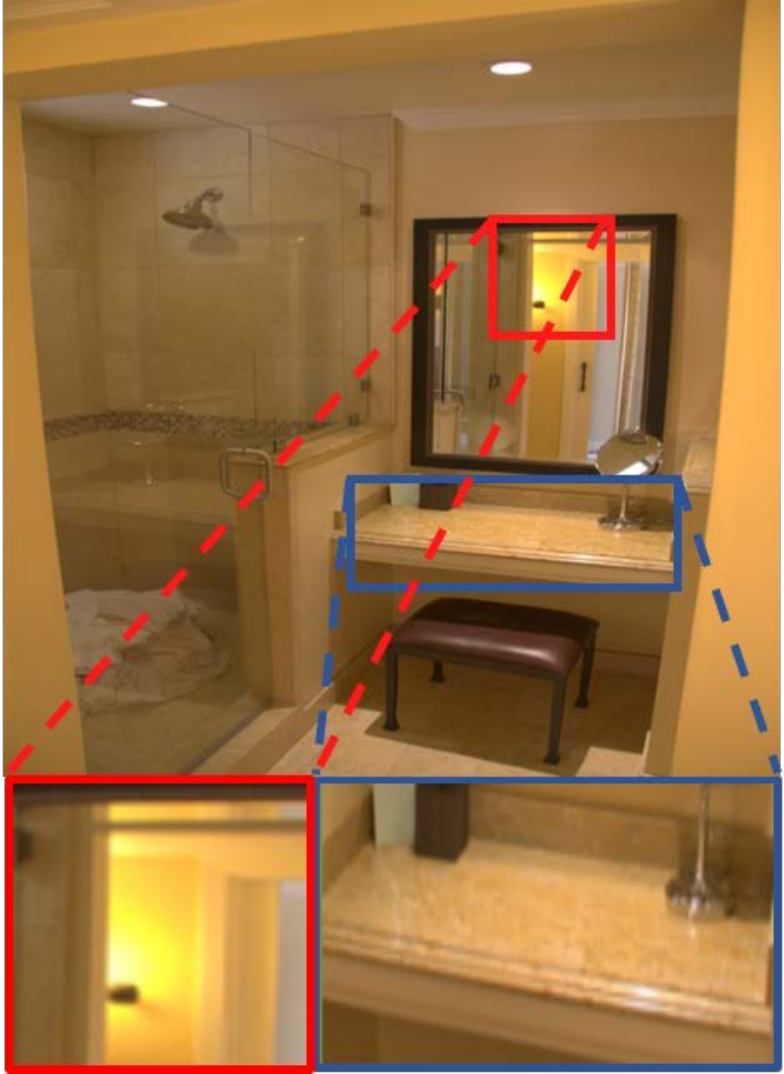}&
			\includegraphics[width=.15\textwidth,height=3.5cm]{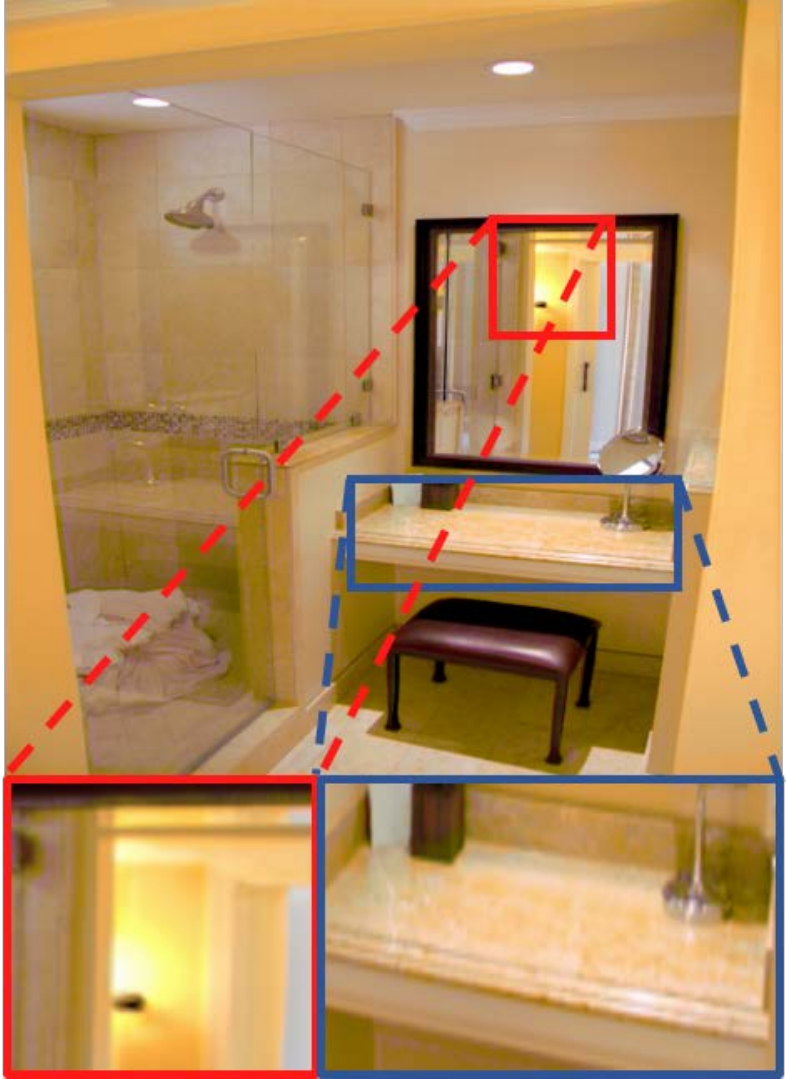}&
			\includegraphics[width=.15\textwidth,height=3.5cm]{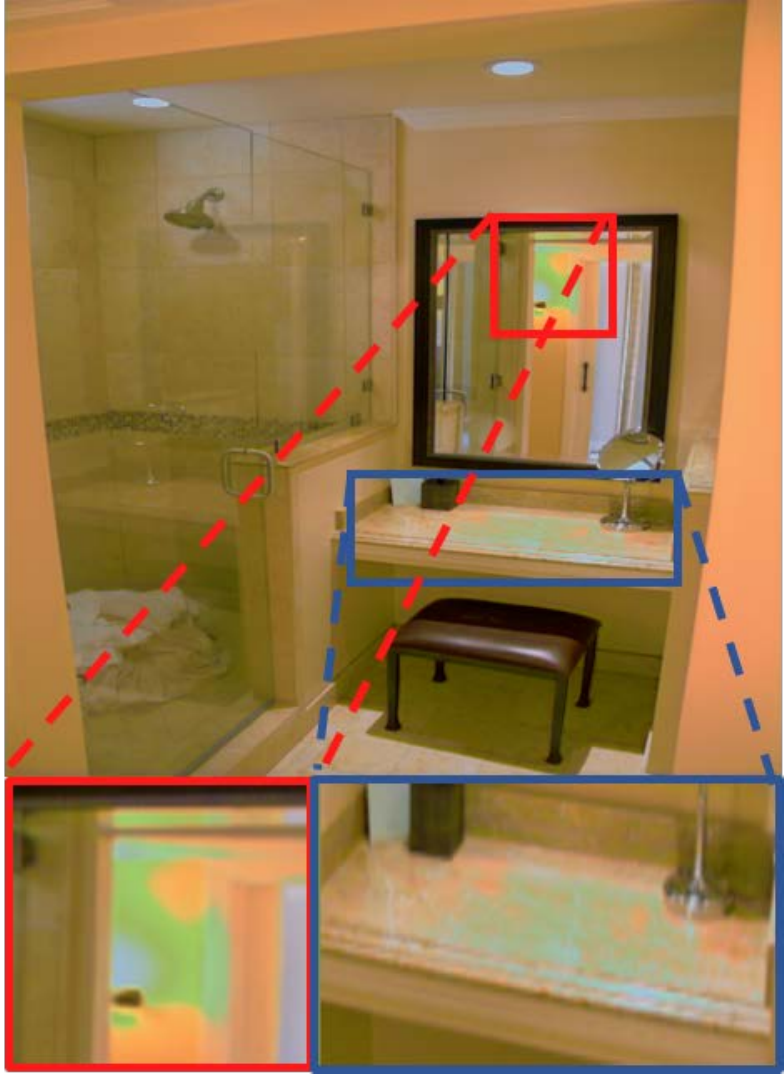}\\
			(a) input image & (b) FlexiCurve-unpair & (c) -w/o LD \\
		\end{tabular}
	\end{center}
	\caption{A visual comparison between  FlexiCurve-unpair and FlexiCurve-unpair-w/o LD (without local discriminator). Red and blue boxes indicate the amplified details.}
	\label{fig:LD}
\end{figure}

\noindent
\textbf{Effect of Input Image Sizes.}
We analyze the effects of input image size on the enhancement performance and the cost of computational resources in our framework. This ablation study is carried on Adobe 5K-T testing set and quantitative results are presented in Table \ref{table_inputsize}. Specifically, we first downsample the size of input image and then feed different sizes of input image to the FlexiCurve-pair. After estimating curve parameters and confidence maps, we upsample the confidence maps back to the same size as the original resolution of input image. With the upsampled confidence maps and curve parameters, we achieve the final result by the plain SCM Fusion.

As shown in Table \ref{table_inputsize}, 2$\times$ downsampling the size of input image has an unnoticeable effect on the enhancement performance (a decline from 23.97dB to 23.26dB, measured in PSNR) but significantly saves computational resources (reduce from 21.97G to 5.49G, measured in FLOPs). With the increase of the times of downsampling, the enhancement performance decreases, but the cost of computational resources is reduced. For example, 4$\times$ downsampling input image only costs 1.37G FLOPs, which leads to a tiny deep model and is suitable for resource-limited devices. As shown, our method can provide a way to balance the enhancement performance and the cost of  computational resources  by feeding different sizes of input images  to the framework.

\begin{table*}
	\centering
	\caption{Quantitative comparisons in terms of PSNR (in dB), SSIM, and LPIPS on Adobe 5K-T testing set, and the trainable parameters (\#P, in M), FLOPs (in G), and runtime (RT, in second). `-' indicate the result is unavailable. The best result is in red whereas the second best one and the third best one are in blue and cyan under each case, respectively.}
	\begin{tabular}{c|c|c|c|c|c|c}
		\hline
		\textbf{Method} & \textbf{PSNR$\uparrow$ } & \textbf{SSIM$\uparrow$} &  \textbf{LPIPS$\downarrow$}& \textbf{\#P$\downarrow$}  &  \textbf{FLOPs$\downarrow$}& \textbf{RT$\downarrow$}\\
		\hline
		input  image         &   17.82     & 0.78 &0.15 & -&- &- \\
		CLAHE \cite{CLAHE}        & 16.71       & 0.78&0.17 & -&- &1.781\\
		PS Auto          &   20.16    & 0.80   &  0.13   &   -&- &-\\
		FCN \cite{Chen17}          & 16.55 & 0.74 &0.20 &  {\color{cyan}0.75} &{\color{red}19.66}&0.038\\
		DPE \cite{Enhancer} & {\color{cyan}23.80}  & {\color{blue}0.90} & {\color{blue}0.07}&   3.35 &{\color{blue}20.75} &{\color{blue}0.010}\\
		DeepLPF \cite{Moran20} &{\color{blue}23.93} &{\color{blue}0.90} &- &1.80 &- &- \\
		DPED\_Iphone \cite{DSLR}      &   18.93    &0.72 & 0.27&  {\color{blue}0.40} &210.91 &0.043\\
		DPED\_Sony \cite{DSLR}      & 19.58     & 0.72 & 0.22 &{\color{blue}0.40} &210.91 &0.043\\
		Exposure \cite{WhiteBox}   & 19.21      &0.82  &0.14  &8.56  &- &4.706\\
		EnlightenGAN \cite{Jiang2019}   & 13.26      &0.74   &0.17  &8.64  &65.84 &{\color{red}0.007}\\
		CycleGAN \cite{CycleGANs}   &  20.73      & 0.78 & 0.27 & 11.38 & 207.87&0.039\\
		FlexiCurve-unpair   &  22.12     & {\color{cyan}0.86} & {\color{cyan}0.11} &{\color{red}0.15} & {\color{cyan}21.97} & {\color{cyan}0.012}\\
		FlexiCurve-pair     & {\color{red}23.97}   &{\color{red}0.91}  & {\color{red}0.06} & {\color{red}0.15} & {\color{cyan}21.97}& {\color{cyan}0.012}\\
		\hline
	\end{tabular}
	\label{table3}
\end{table*}

\noindent
\textbf{Benefit of Spatial Interpolation.}
We study the benefit of spatial interpolation. In addition to Figure \ref{fig:user}, we show another example with the amplified details in Figure \ref{fig:si}, which follows the same experimental settings. As shown, the automatic result has enhanced contrast and saturation, however, some local regions such as the region in the red box in Figure \ref{fig:si}(b) still suffer from insufficient lighting and unpleasing color. The spatial interpolation is capable of adjusting the local regions, thus achieving sufficient brightness and natural color. In this way, the users can obtain more visually pleasing and diverse results.

\noindent
\textbf{Contribution of Local Discriminator.}
For FlexiCurve with unpaired data, we add a local discriminator to improve the local details of the final result. `FlexiCurve-unpair-w/o LD' corresponds to the ablated model that only employs a global discriminator.  A visual comparison between FlexiCurve-unpair and FlexiCurve-unpair-w/o LD is shown in Figure \ref{fig:LD}. As shown, the result of FlexiCurve-unpair-w/o LD introduces  greenish color in local regions as indicated in red and blue boxes since the global discriminator alone is often unable to handle the local details well.

\begin{figure*}
	\begin{center}
		\begin{tabular}{c@{ }c@{ }c@{ }c@{ }c@{ }c@{ }c@{ }}
			\includegraphics[height=5.7cm,width=2.8cm]{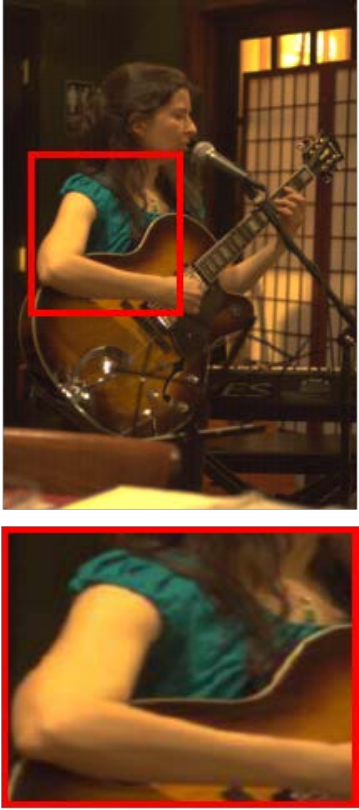}&
			\includegraphics[height=5.7cm,width=2.8cm]{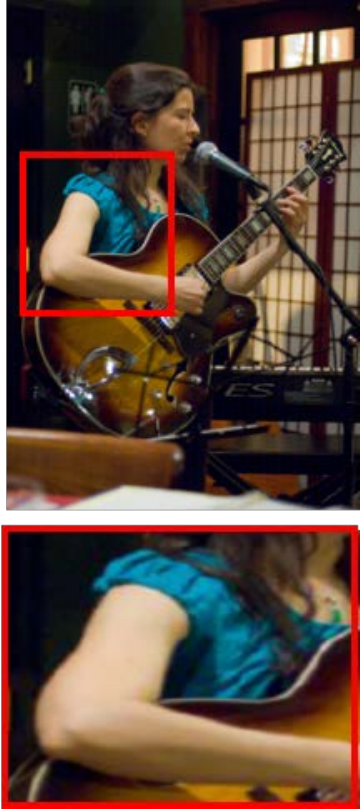}&
			\includegraphics[height=5.7cm,width=2.8cm]{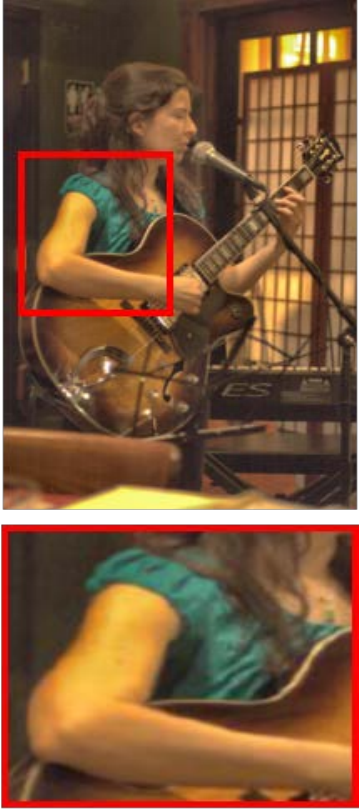}&
			\includegraphics[height=5.7cm,width=2.8cm]{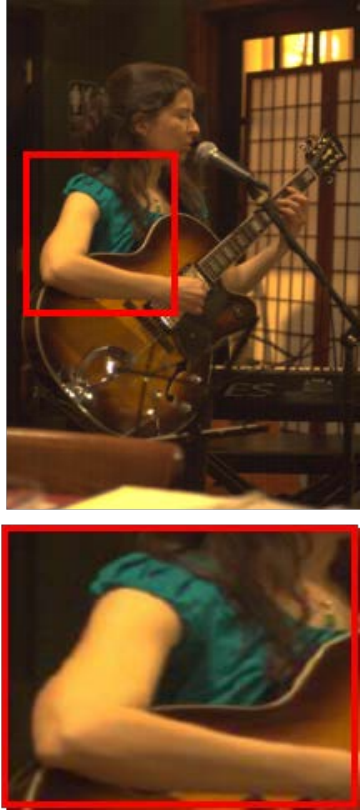}&
			\includegraphics[height=5.7cm,width=2.8cm]{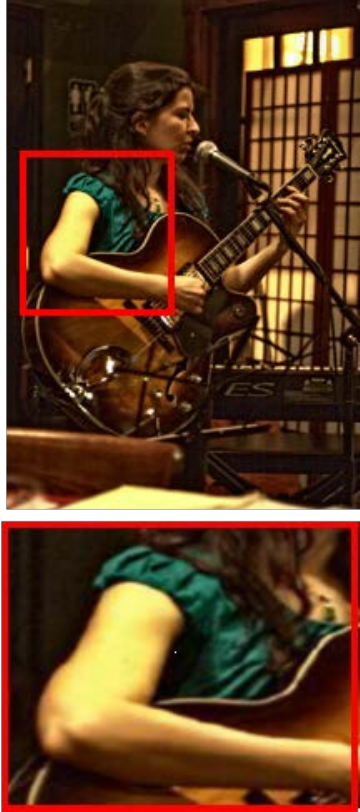}&
			\includegraphics[height=5.7cm,width=2.8cm]{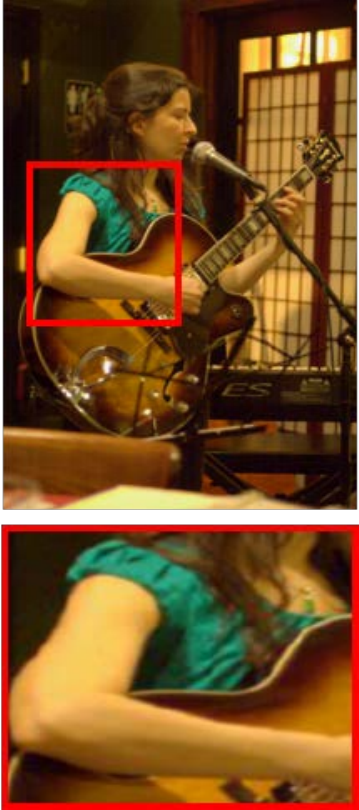} \\
			(a) input image &  (b) Expert C  & (c)  CLAHE \cite{CLAHE}  &(d) PS Auto &(e) FCN \cite{Chen17} & (f) DPE \cite{Enhancer} \\
			\includegraphics[height=5.7cm,width=2.8cm]{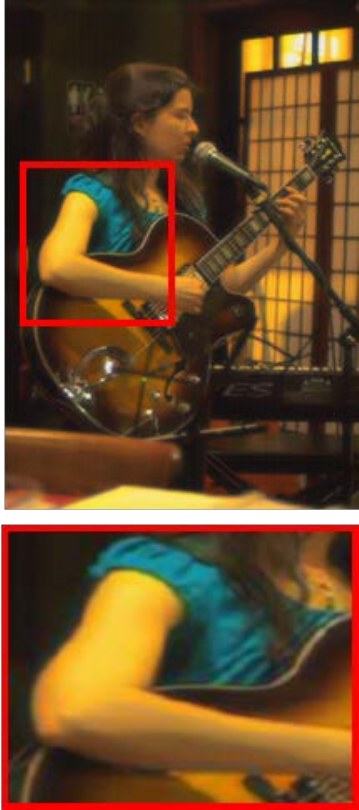}&
			\includegraphics[height=5.7cm,width=2.8cm]{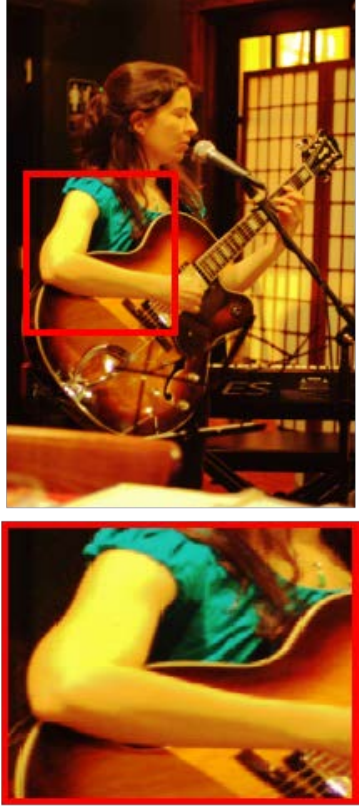}&
			\includegraphics[height=5.7cm,width=2.8cm]{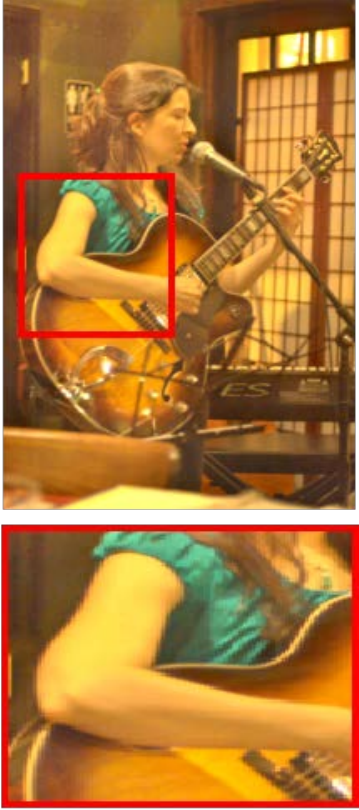}&
			\includegraphics[height=5.7cm,width=2.8cm]{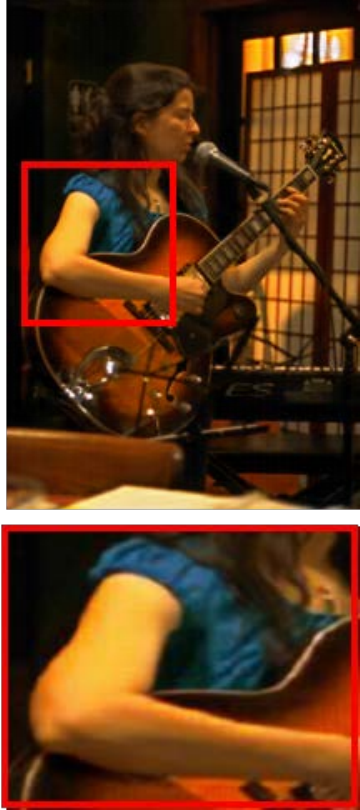}&
			\includegraphics[height=5.7cm,width=2.8cm]{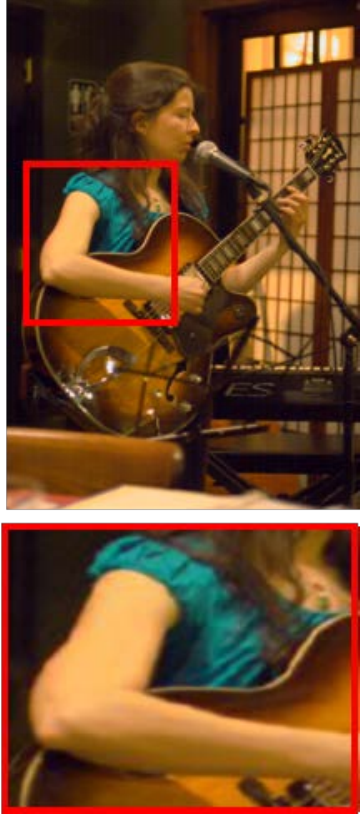}&
			\includegraphics[height=5.7cm,width=2.8cm]{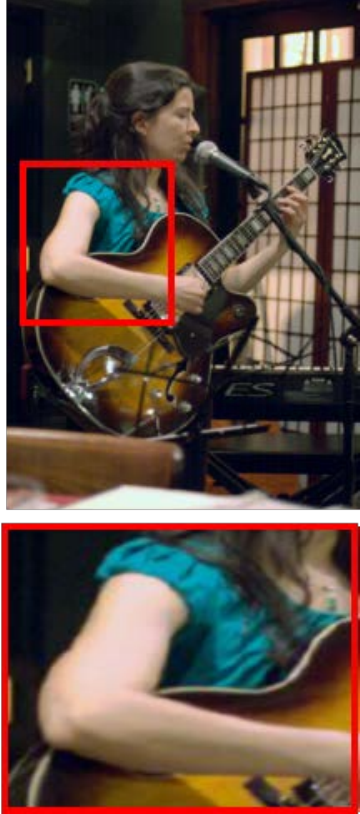} \\
			(g) DPED\_Sony \cite{DSLR}  & (h) Exposure \cite{WhiteBox}& (i) EnlightenGAN \cite{Jiang2019} & (j) CycleGAN \cite{CycleGANs}  &(k) FlexiCurve-unpair &(l) FlexiCurve-pair \\
		\end{tabular}
	\end{center}
	\caption{Visual comparisons on typical low-quality image sampled from Adobe 5K-T testing set. Red boxes indicate amplified details.}
	\label{fig:visual_5K1}
\end{figure*}

\begin{figure*}
	\begin{center}
		\begin{tabular}{c@{ }c@{ }c@{ }c@{ }c@{ }c@{ }c@{ }}
			\includegraphics[height=4.2cm,width=2.8cm]{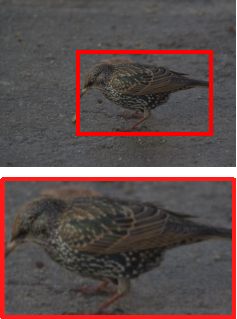}&
			\includegraphics[height=4.2cm,width=2.8cm]{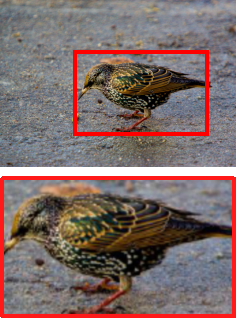}&
			\includegraphics[height=4.2cm,width=2.8cm]{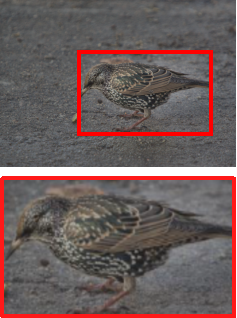}&
			\includegraphics[height=4.2cm,width=2.8cm]{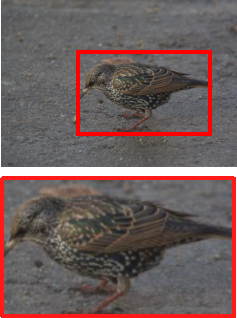}&
			\includegraphics[height=4.2cm,width=2.8cm]{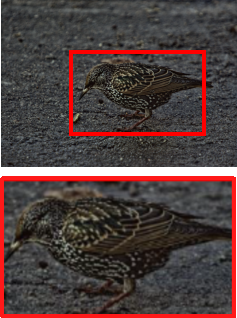} &
			\includegraphics[height=4.2cm,width=2.8cm]{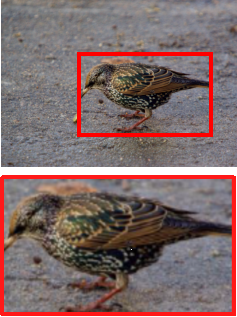}\\
			(a) input image &  (b) Expert C  & (c)  CLAHE \cite{CLAHE}  &(d) PS Auto &(e) FCN \cite{Chen17}& (f) DPE \cite{Enhancer} \\
			\includegraphics[height=4.2cm,width=2.8cm]{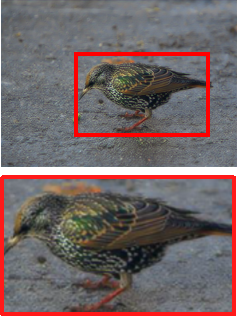}&
			\includegraphics[height=4.2cm,width=2.8cm]{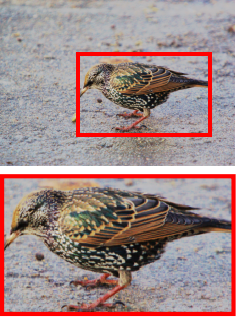}&			
			\includegraphics[height=4.2cm,width=2.8cm]{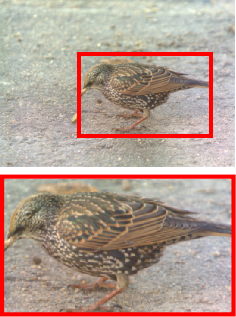}&
			\includegraphics[height=4.2cm,width=2.8cm]{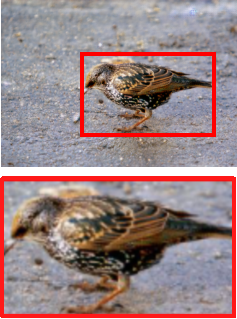}&
			\includegraphics[height=4.2cm,width=2.8cm]{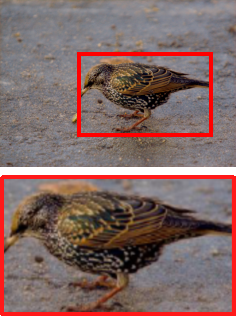}&
			\includegraphics[height=4.2cm,width=2.8cm]{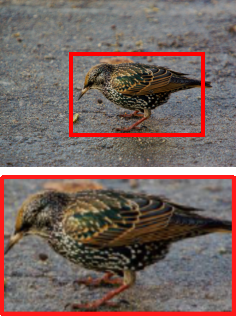} \\
			(g) DPED\_Sony \cite{DSLR} & (h) Exposure \cite{WhiteBox}& (i) EnlightenGAN \cite{Jiang2019} & (j) CycleGAN \cite{CycleGANs}  &(k) FlexiCurve-unpair &(l) FlexiCurve-pair  \\
		\end{tabular}
	\end{center}
	\caption{Visual comparisons on typical low-quality image sampled from Adobe 5K-T testing set. Red boxes indicate amplified details.}
	\label{fig:visual_5K3}
\end{figure*}

\begin{figure*}
	\begin{center}
		\begin{tabular}{c@{ }c@{ }c@{ }c@{ }c@{ }c@{ }c@{ }}
			\includegraphics[height=4.2cm,width=2.8cm]{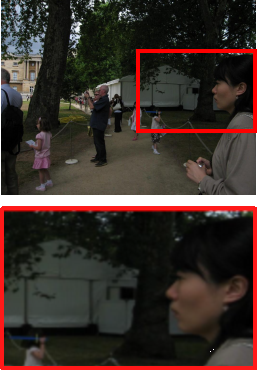}&
			\includegraphics[height=4.2cm,width=2.8cm]{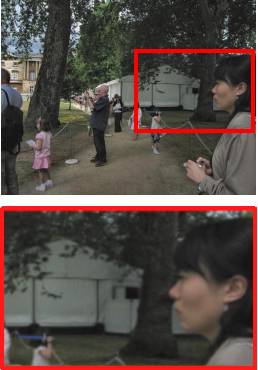}&
			\includegraphics[height=4.2cm,width=2.8cm]{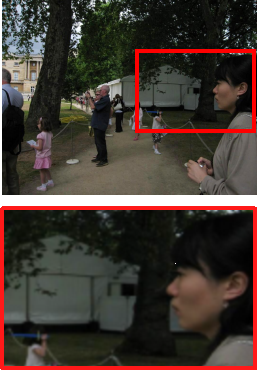}&
			\includegraphics[height=4.2cm,width=2.8cm]{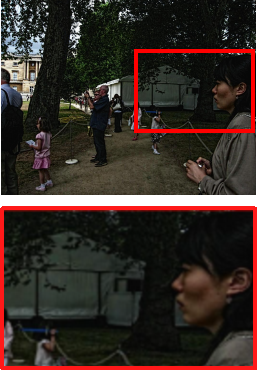}&
			\includegraphics[height=4.2cm,width=2.8cm]{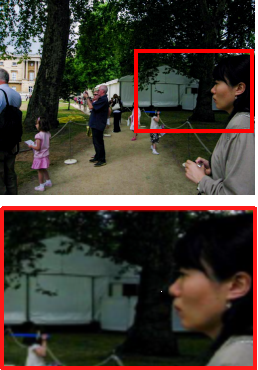} &
			\includegraphics[height=4.2cm,width=2.8cm]{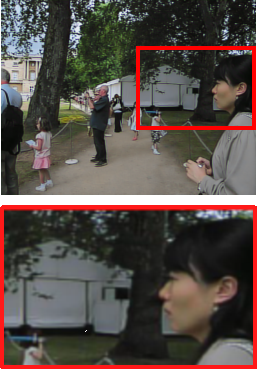}\\
			(a) input image &  (b)  CLAHE \cite{CLAHE}  &(c) PS Auto &(d) FCN \cite{Chen17} &(e) DPE \cite{Enhancer} & (f) DPED\_Iphone \cite{DSLR} \\
			\includegraphics[height=4.2cm,width=2.8cm]{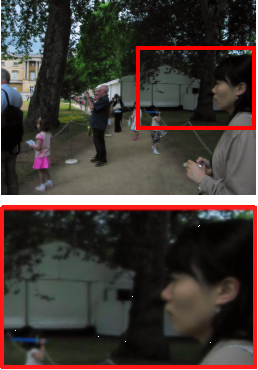}&
			\includegraphics[height=4.2cm,width=2.8cm]{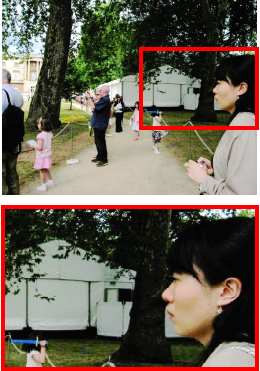}&
			\includegraphics[height=4.2cm,width=2.8cm]{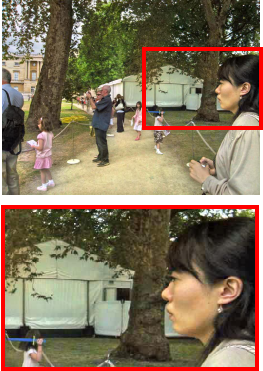}&			
			\includegraphics[height=4.2cm,width=2.8cm]{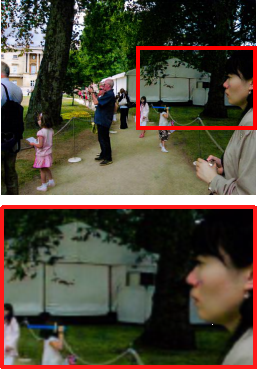}&
			\includegraphics[height=4.2cm,width=2.8cm]{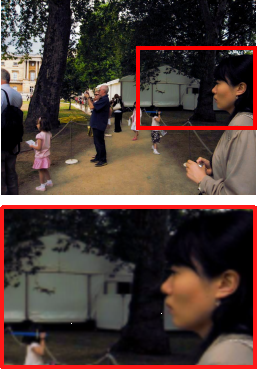}&
			\includegraphics[height=4.2cm,width=2.8cm]{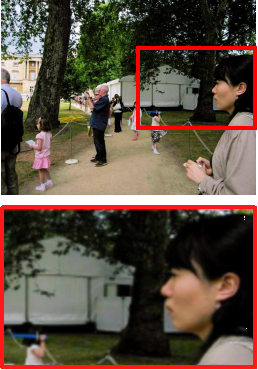} \\
			(g) DPED\_Sony \cite{DSLR}   & (h) Exposure \cite{WhiteBox}& (i) EnlightenGAN \cite{Jiang2019} & (j) CycleGAN \cite{CycleGANs}  &(k) FlexiCurve-unpair &(l) FlexiCurve-pair \\
		\end{tabular}
	\end{center}
	\caption{Visual comparisons on low-quality image sampled from CUHK-PQ testing set. Red boxes indicate amplified details.}
	\label{fig:visual_CUHK}
\end{figure*}

\subsection{Perceptual Comparisons}
We present the visual comparisons on typical low-quality images sampled from Adobe 5K-T testing sets in Figures~\ref{fig:visual_5K1} and ~\ref{fig:visual_5K3}. In Figure~\ref{fig:visual_5K1}, all methods fail to restore the color of input images such as the color of clothing and skin, except for FlexiCurve-pair. 
FCN \cite{Chen17} and Exposure \cite{WhiteBox} introduce artifacts while CLAHE \cite{CLAHE}, DPED\_Sony \cite{DSLR}, and CycleGAN \cite{CycleGANs} produce blurring details. EnlightenGAN \cite{Jiang2019} over-enhances local regions.
As presented in Figure~\ref{fig:visual_5K3}, our FlexiCurve-pair achieves good contrast and clear detail in spite of its lightweight network structure. Although Exposure \cite{WhiteBox} improves the contrast and brightness, it tends to introduce over-exposed  regions such as the ground regions. The methods of CLAHE  \cite{CLAHE}, PS Auto, and EnlightenGAN \cite{Jiang2019} degrade the contrast of input image and introduce a haze-like appearance. FCN \cite{Chen17} generates obvious artifacts. In contrast, FlexiCurve-unpair effectively enhances the input low-quality image without introducing obvious blurring unlike the result of CycleGAN \cite{CycleGANs} that blurs the detail. 

To validate the generalization capability of our method, we show the visual comparisons on low-quality image sampled from CUHK-PQ testing set in Figure~\ref{fig:visual_CUHK}. CLAHE \cite{CLAHE} and PS Auto cannot effectively enhance the input image while DPED\_Sony has less effect on the input image. FCN \cite{Chen17} produces artifacts such as the region of the face. Exposure \cite{WhiteBox} over-enhances the face region. CycleGAN \cite{CycleGANs} introduces the artificial color such as the red nose. EnlightenGAN \cite{Jiang2019} produces color deviations such as the color of leaves. In summary, the competing methods introduce artifacts, blurring or unpleasing color while both FlexiCurve-pair and FlexiCurve-unpair not only produce visually pleasing results, but also effectively improve the brightness of input image.

\subsection{Quantitative Comparisons}
We carry out quantitative comparisons by using the full-reference IQA metrics PSNR, SSIM~\cite{SSIM}, and LPIPS \cite{LPIPS} on the Adobe 5K-T testing set and also report the trainable parameters, FLOPs, and runtime of different methods. We use the same codes as \cite{Enhancer} to compute the PSNR and SSIM values. For the evaluation metric LPIPS \cite{LPIPS}, we employ the AlexNet-based model to compute the perceptual  similarity based on its efficiency. A lower LPIPS value suggests a result is closer to the ground truth in terms of perceptual similarity. We compare the FLOPs and runtime of different methods on an image with a size of 512$\times$512$\times$3.
For FCN \cite{Chen17}, we compute the parameters and FLOPs by using its CAN32+AN model.
For CycleGAN \cite{CycleGANs}, we only compute the network trainable parameters, FLOPs, and runtime of its generator that produces the final enhanced result. 
For Exposure \cite{WhiteBox}, we did not consider its FLOPs because the retouching operations are variational based on the decision of reinforcement learning in its framework.

In Table~\ref{table3}, our FlexiCurve-pair achieves the best average values of PSNR, SSIM and LPIPS, which indicates the good content, structure, and perceptual similarity between our results and the ground truth. The performance of our FlexiCurve-pair is even superior to the pixel-wise reconstruction-based method DPE \cite{Enhancer} that comes with a large network. In addition, our FlexiCurve-unpair obtains better quantitative performance than the unpaired EnlightenGAN \cite{Jiang2019} and CycleGAN \cite{CycleGANs}, which further demonstrates the effectiveness of FlexiCurve framework regardless of paired or unpaired training data. 

As presented in Table~\ref{table3}, our method also achieves the smallest network parameters, which is about 2.6 times smaller than the second-smallest deep model DPED \cite{DSLR} and 70 times smaller than the largest one CycleGAN \cite{CycleGANs}. Such a lightweight network of FlexiCurve is desired by the resources-limited platforms.  In terms of the FLOPs and runtime, our method also achieves comparable performance with the best performer, but has advantages on enhancement performance and flexibility over these methods. Besides, our method can significantly reduce the cost of FLOPs by downsampling the size of input image when compromising the enhancement performance as shown in Table \ref{table_inputsize}. However, the enhancement performance (measured in PSNR) of the downsampling input image is still better than most of the  compared methods. In summary, in spite of employing a plain and lightweight network structure, our method can achieve the best performance across perceptual and quantitative comparisons, which benefits from the unique design of global piecewise curve adjustment coupled with spatial-adaptive fusion.

\begin{table}
	\centering
	\caption{User study on CUHK-PQ testing set. The best value is in red.}
	\begin{tabular}{c|c|c|c}
		\hline
		\textbf{Method} & 	\textbf{Score} &\textbf{Method}  & 	\textbf{Score}  \\
		\hline
		input image & 0 &  DPED\_Sony \cite{DSLR} &  8$\%$ \\
		CLAHE \cite{CLAHE} & 0 &   Exposure \cite{WhiteBox} &   11$\%$\\ 
		PS Auto & 2$\%$ & EnlightenGAN \cite{Jiang2019} & 4$\%$ \\ 
		FCN \cite{Chen17} & 2$\%$ & CycleGAN \cite{CycleGANs} &  10$\%$\\ 
		DPE \cite{Enhancer}  & 22$\%$ & FlexiCurve-unpair &12$\%$ \\ 
		DPED\_Iphone \cite{DSLR}  &6$\%$&  FlexiCurve-pair & {\color{red}25$\%$} \\
		\hline
	\end{tabular}
	\label{table4}
\end{table}

\begin{figure*}
	\centering
	\centerline{\includegraphics[width=0.9\textwidth]{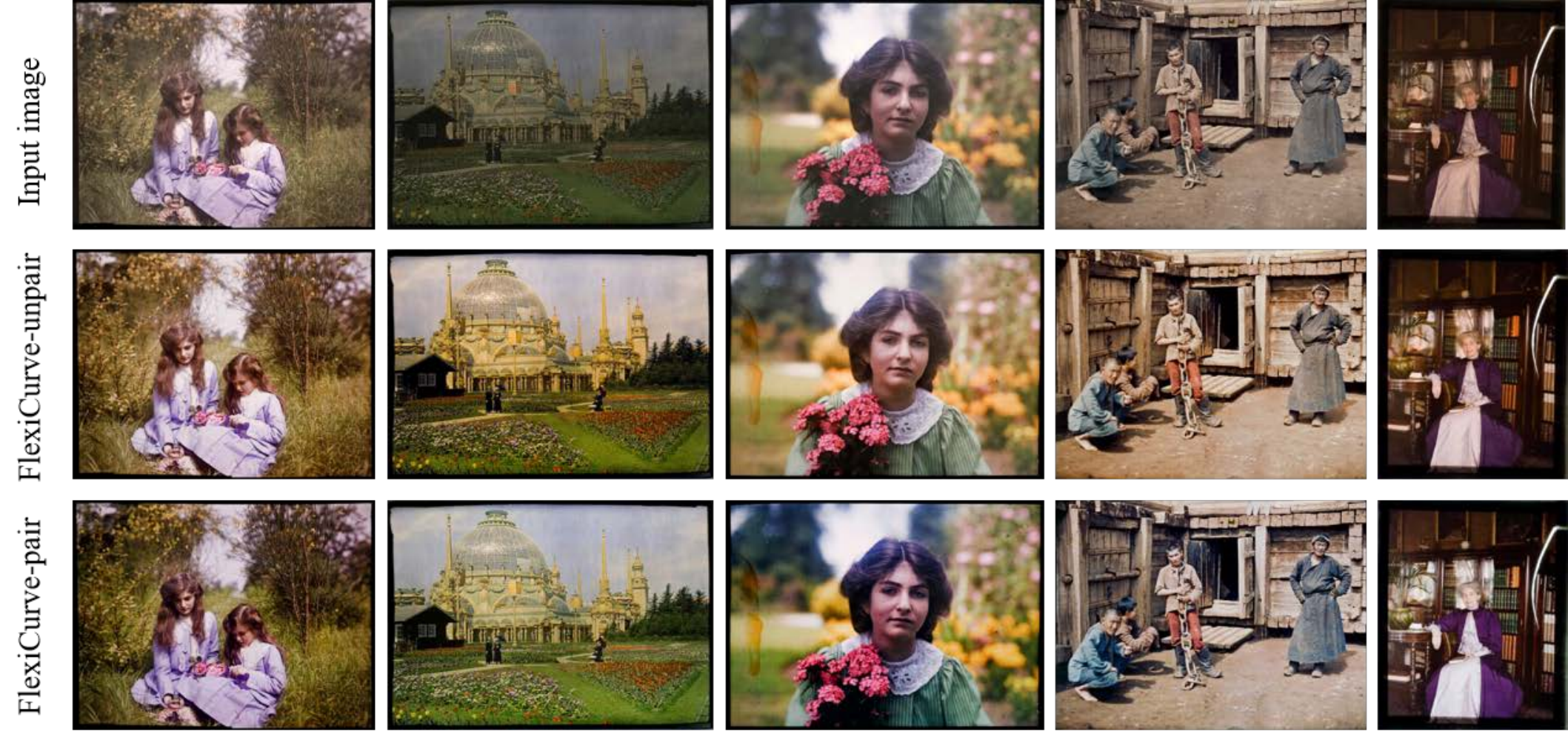}}
	\caption{A set of examples of applying our method to enhance the oldest color photos. Both paired and unpaired framewoks effectively recover the color and improve the contrast and brightness of the old color photos.}
	\label{fig:old_photo}
\end{figure*}

\subsection{User Study}
We perform a user study to measure the subjective visual quality of compared methods on CUHK-PQ testing set. We invite 10 participants to independently select the most favorite result  from  each set of enhanced results by various methods for a given input image (each set consists of an input image and the corresponding results by 11  methods). For each set of comparisons, we select the method that produces the result selected by most subjects as the winner. To avoid subjective bias, we shuffle the results of different methods. We totally conduct 80 sets of comparisons. If the results achieve the same number of votes, we will treat all as the winner for this image. These participants are trained by observing the results from 1) whether the results contain over-/under-exposed artifacts or over-/under-enhanced regions; 2) whether the results introduce artificial colors; 3) whether the results have  unnatural texture and obvious noise; and 4) whether the results have inappropriate contrast. We present the percentage of the results is selected in Table \ref{table4}. 

Compared with various methods, our FlexiCurve-pair is  most favored by the participants, reaching 25$\%$. This result suggests that FlexiCurve-pair can produce visually pleasing results and obtain good generalization capability. Moreover, both our FlexiCurve-unpair  and CycleGAN \cite{CycleGANs} are selected frequently (12$\%$ and 10$\%$, respectively) thanks to the adversarial learning used in unpaired training. This is because that the GAN-based curve estimation manner can effectively avoid the risk of overfitting on specific data. 

\subsection{Addition Results}
\label{Application}
To further demonstrate the generalization capability of our method, we apply our method to enhance the oldest color photos in the world and show the results in Figure \ref{fig:old_photo}. 
As shown in Figure \ref{fig:old_photo}, both paired and unpaired frameworks effectively refresh the old color photos in terms of color, contrast, saturation, and brightness.

\section{Conclusion and Discussion}
\label{sec:Limitation and Discussion}

We proposed a novel curve estimation method for photo enhancement, called FlexiCurve. Photos can be enhanced via specially designed piecewise curves coupled with spatial-adaptive fusion. The estimation of curve parameters and confidence maps are achieved within a multi-task network. The proposed FlexiCurve is not limited to paired or unpaired data. We evaluate the effectiveness of our method on diverse scenes by using various metrics qualitatively and quantitatively. Experiments demonstrate the effectiveness of our method against existing methods. Besides, the lightweight structure, fast inference speed, one-to-many global enhancement, and spatial interpolation also suggest the potential of our method in practical applications.

\begin{figure}[t]
	\begin{center}
		\begin{tabular}{c@{ }c@{ }c@{ }c}
			\includegraphics[width=.15\textwidth,height=2.1cm]{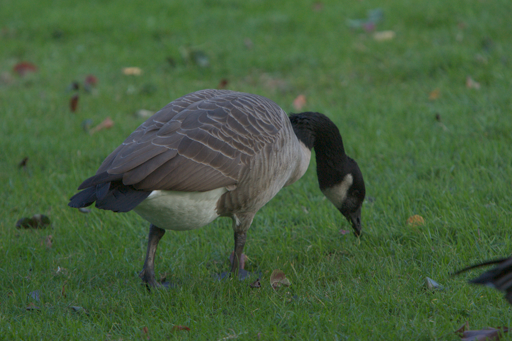}&
			\includegraphics[width=.15\textwidth,height=2.1cm]{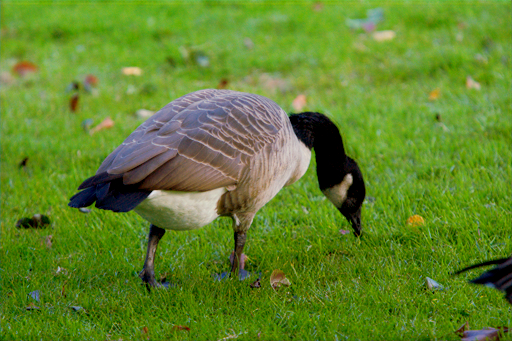}&
			\includegraphics[width=.15\textwidth,height=2.1cm]{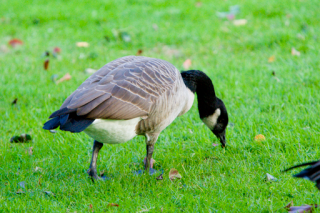}\\
			(a) input image & (b) FlexiCurve-unpair & (c) ground truth \\
		\end{tabular}
	\end{center}
	\caption{Failure case by FlexiCurve-unpair. The color in the result of FlexiCurve-unpair is different from that of the input image and the ground truth.}
	\label{fig:CC}
\end{figure}

Like previous unsupervised models \cite{CycleGANs,unsupervisedGANs}, the proposed FlexiCurve-unpair may suffer from color casts. As shown in Figure \ref{fig:CC}, the colors (such as the green color) in the resulting image of FlexiCurve-unpair are different from that of the input image and the ground truth. This is mainly because of the domain gap between source image collection and target image collection in the process of unsupervised learning.
In addition, since our spatial interpolation highly depends on the confidence maps and is controlled by three adjustment weights $\alpha$, $\beta$, and $\gamma$, all pixels with the same values in a confidence map will be adjusted identically regardless of their spatial positions. Thus, our spatial interpolation cannot interpolate arbitrary local regions while keeping other regions invariant.

\ifCLASSOPTIONcaptionsoff
\newpage
\fi

{
	\bibliographystyle{IEEEtran}
	\bibliography{bibliography}
}
\begin{IEEEbiography}[{\includegraphics[width=1in,height=1.25in,clip,keepaspectratio]{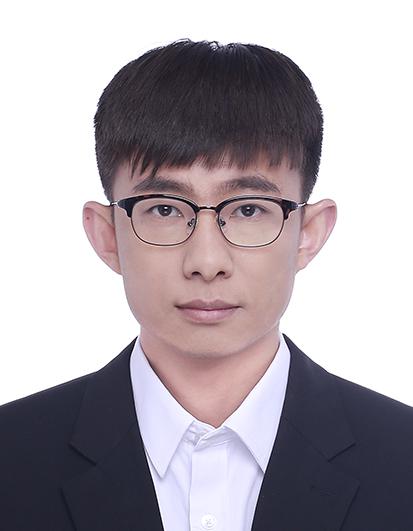}}]{Chongyi Li}  received the Ph.D. degree from the School of Electrical and Information Engineering, Tianjin University, Tianjin, China, in June 2018. From 2016 to 2017, he was a joint-training Ph.D. Student with Australian National University, Australia. He was a postdoctoral fellow with the Department of Computer Science, City University of Hong Kong, Hong Kong. He is currently a research fellow with the School of Computer Science and Engineering, Nanyang Technological University (NTU), Singapore. His current research focuses on image processing, computer vision, and deep learning, particularly in the domains of image restoration and enhancement.
\end{IEEEbiography}

\begin{IEEEbiography}[{\includegraphics[width=1in,height=1.25in,clip,keepaspectratio]{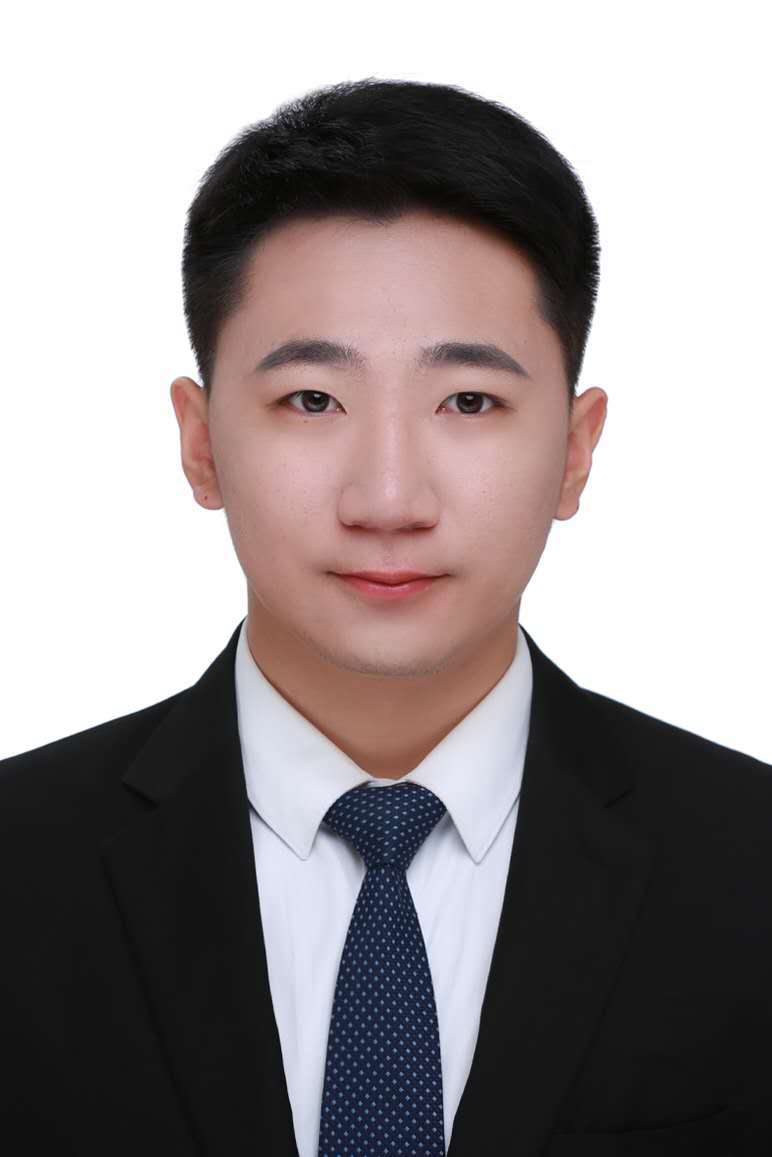}}]{Chunle Guo}  received his PhD degree from Tianjin University in China under the supervision of Prof. Jichang Guo. He conducted the Ph.D. research as a Visiting Student with the School of Electronic Engineering and Computer Science, Queen Mary University of London (QMUL), UK.  He continued his research as a Research Associate with Department of Computer Science, City University of Hong Kong (CityU), from 2018 to 2019. Now he is a postdoc research fellow working with Prof. Ming-Ming Cheng in Nankai University. His research interests lies in image processing, computer vision, and deep learning.
\end{IEEEbiography}
\begin{IEEEbiography}[{\includegraphics[width=1in,height=1.25in,clip,keepaspectratio]{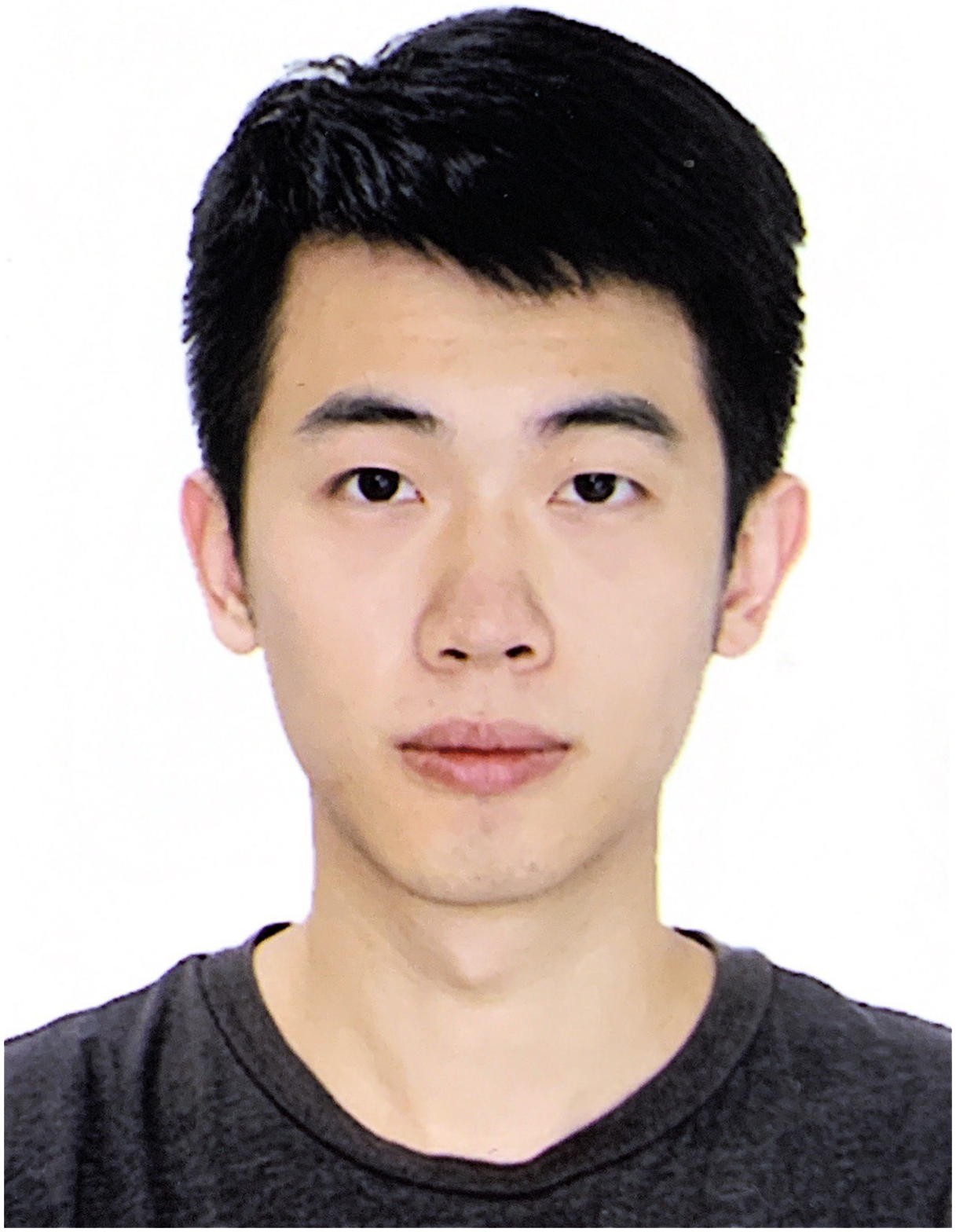}}]{Qiming Ai} is currently pursuing his master’s degree at Nanyang Technological University in Singapore. Previously, he received his B.Eng. degree from the University of Science and Technology of China and served as a research engineer with the School of Computer Science and Engineering, Nanyang Technological University, SG. His research interests lie in computer vision and multimodality (vision and language).
\end{IEEEbiography}
\begin{IEEEbiography}[{\includegraphics[width=1in,height=1.25in,clip,keepaspectratio]{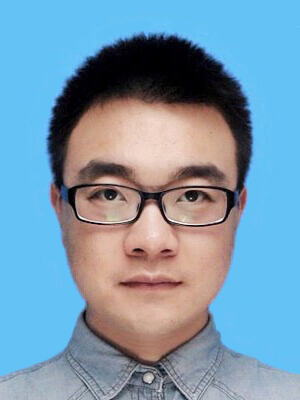}}]{Shangchen Zhou} is currently a Ph.D. student at Nanyang Technological University. Before that, he received his B.Eng. and M.Eng. degrees from the University of Electronic Science and Technology of China and Harbin Institute of Technology in 2015 and 2018, respectively. His research interests include computer vision and image processing.
\end{IEEEbiography}
\begin{IEEEbiography}[{\includegraphics[width=1in,height=1.25in,clip,keepaspectratio]{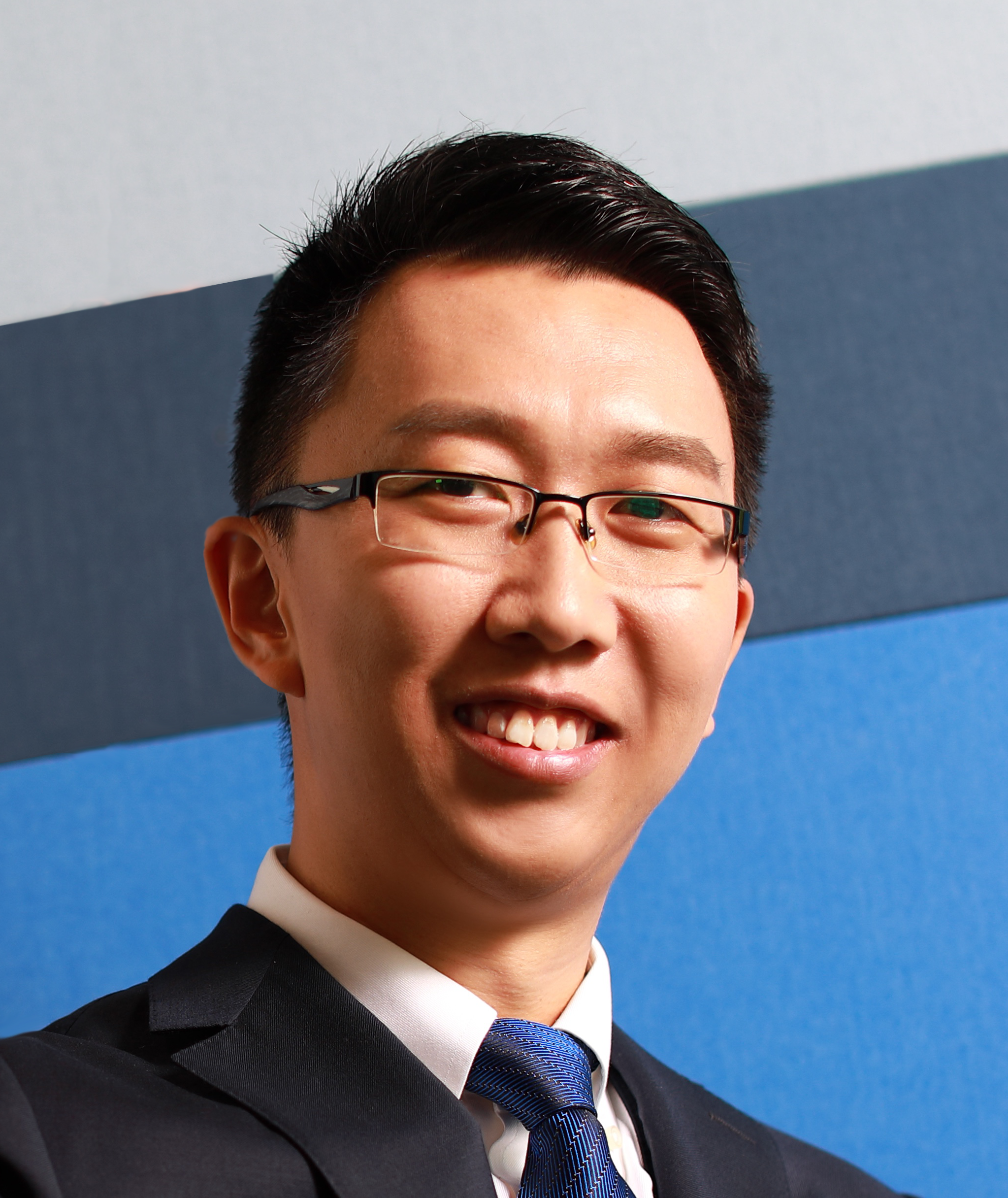}}]{Chen Change Loy} (Senior Member, IEEE) received the PhD degree in computer science from the Queen Mary University of London, in 2010. He is an associate professor with the School of Computer Science and Engineering, Nanyang Technological University. Prior to joining NTU, he served as a research assistant professor with the Department of Information Engineering, The Chinese University of Hong Kong, from 2013 to 2018. His research interests include computer vision and deep learning. He serves as an associate editor of the IEEE Transactions on Pattern Analysis and Machine Intelligence and the International Journal of Computer Vision. He also serves/served as an Area Chair of CVPR 2021, CVPR 2019, ECCV 2018, AAAI 2021 and BMVC 2018-2020.
\end{IEEEbiography}

\end{document}